\documentclass[runningheads]{llncs}
\usepackage{graphicx}
\usepackage{comment}
\usepackage{amsmath,amssymb} 
\pdfoutput=1
\usepackage{color}
\usepackage{colortbl}
\usepackage{float}
\usepackage{lipsum}
\usepackage{cite}
\usepackage[title]{appendix}
\usepackage{multirow}
\usepackage{xfrac}
\usepackage[width=122mm,left=12mm,paperwidth=146mm,height=193mm,top=12mm,paperheight=217mm]{geometry}

\definecolor{myRed}{rgb}{0.8,0.2,0.2}
\definecolor{myBlue}{rgb}{0.2,0.2,0.8}

\newcommand{\etal}{\textit{et al.}}
\newcommand{\tabincell}[2]{\begin{tabular}{@{}#1@{}}#2\end{tabular}}
\begin{document}
\pagestyle{headings}
\mainmatter

\title{SlimConv: Reducing Channel Redundancy in Convolutional Neural Networks by Weights Flipping} 


\titlerunning{SlimConv}
%
\author{Jiaxiong Qiu\and
Cai Chen\and
Shuaicheng Liu\thanks{indicates corresponding author.}\and 
Bing Zeng}
%
%
\institute{University of Electronic Science and Technology of China}
\maketitle

\begin{abstract}
The channel redundancy in feature maps of convolutional neural networks (CNNs) results in the large consumption of memories and computational resources. In this work, we design a novel Slim Convolution (SlimConv) module to boost the performance of CNNs by reducing channel redundancies. Our SlimConv consists of three main steps: \textit{Reconstruct}, \textit{Transform} and \textit{Fuse}, through which the features are splitted and reorganized in a more efficient way, such that the learned weights can be compressed effectively. In particular, the core of our model is a weight flipping operation which can largely improve the feature diversities, contributing to the performance crucially. Our SlimConv is a plug-and-play architectural unit which can be used to replace convolutional layers in CNNs directly. We validate the effectiveness of SlimConv by conducting comprehensive experiments on ImageNet, MS COCO2014, Pascal VOC2012 segmentation, and Pascal VOC2007 detection datasets. The experiments show that SlimConv-equipped models can achieve better performances consistently, less consumption of memory and computation resources than non-equipped conterparts. For example, the ResNet-101 fitted with SlimConv achieves $77.84\%$ top-1 classification accuracy with 4.87 GFLOPs and 27.96M parameters on ImageNet, which shows almost $0.5\%$ better performance with about 3 GFLOPs and $38\%$ parameters reduced.

\end{abstract}

\section{Introduction}
In most studies of deep learning, convolutional neural networks (CNNs) have been emphasized with attention given to their impactful modeling for various vision tasks, such as image classification~\cite{russakovsky2015imagenet}, object detection~\cite{ren2015faster} and semantic segmentation~\cite{chen2018encoder}. Vanilla convolutional layers are designed to be increasingly deeper and more complicated for the better accuracy, but these models bring massive parameters and floating point operations (FLOPs). Hence, the accuracy and cost tradeoffs are currently ubiquitous in efficient CNN design, especially for the mobile and edged devices (e.g., smartphones, drones and self-driving cars). Over the past several decades, the field has spawned a giddy mix of methods to compress the models while preserving accuracies, such as quantizing~\cite{rastegari2016xnor} and pruning~\cite{han2016dsd, li2016pruning}. Besides, hand-craft or automatic designs have also made appreciable success.

MobileNet~\cite{sandler2018mobilenetv2} exerts depth-wise and point-wise convolutions to build a small network with low latency, which achieves over $71\%$ top1-accuracy with merely $3.5$M parameters and $0.3$ GFLOPs on ImageNet~\cite{russakovsky2015imagenet}. Shufflenet~\cite{ma2018shufflenet} resorts to the channel shuffle operation to improve performance of tiny networks owing to the sufficiently process to the inter-channel information. Neural architecture search (NAS) methods~\cite{zoph2018learning, liu2018progressive} learn to tune network architectures and gain efficient models with high potentials.

These methods are mostly dedicated in explicitly diminishing the redundant parameters and decreasing the computational cost. However, the information loss brought by many of them hampers the effort to performance. Researches have investigated that there are plenty of redundant futures in the network channels, which not only determine the model size but also influence the implicit representation of models. OctConv~\cite{chen2019drop} discussed the effectiveness of dropping the low-frequency part of features by interleaving connections in the convolutional layers, thus reducing the spatial redundancy. Besides, the series of Network Slimming~\cite{yu2018slimmable, yu2019universally, yu2019autoslim} conducted nice explorations to directly prune the redundant channels despite of the difficulty in dealing with the batch normalization. In particular, Autoslim~\cite{yu2019autoslim} achieved the state-of-the-art performance with the help of NAS~\cite{zoph2018learning}.

Based on these observations, our studies in this paper focus on the efficient design of channel reduction operation. We propose a novel convolutional operation, which we term as Slim Convolution (SlimConv), with the purpose of obtaining good performances while saving the computational resources simultaneously. The SilmConv can reduce and reform future channels to improve the quality of feature representations. Specifically, the SlimConv is designed as a plug-and-play module which can be embedded into various popular CNN models with only one hyperparameter. The SlimConv not only reduces calculations but also maintains the capability of feature representations, within which a novel operation, weights flipping, plays an important role. It is a light-weight operation, but can maintain representations significantly during the channel reduction. In addition, we further incorporate the attention mechanism into the channel reduction which enables the network to concentrate on the emphasized feature channels while neglecting those of lower weights.

\begin{figure}[t]
\centering
\includegraphics[scale=0.45]{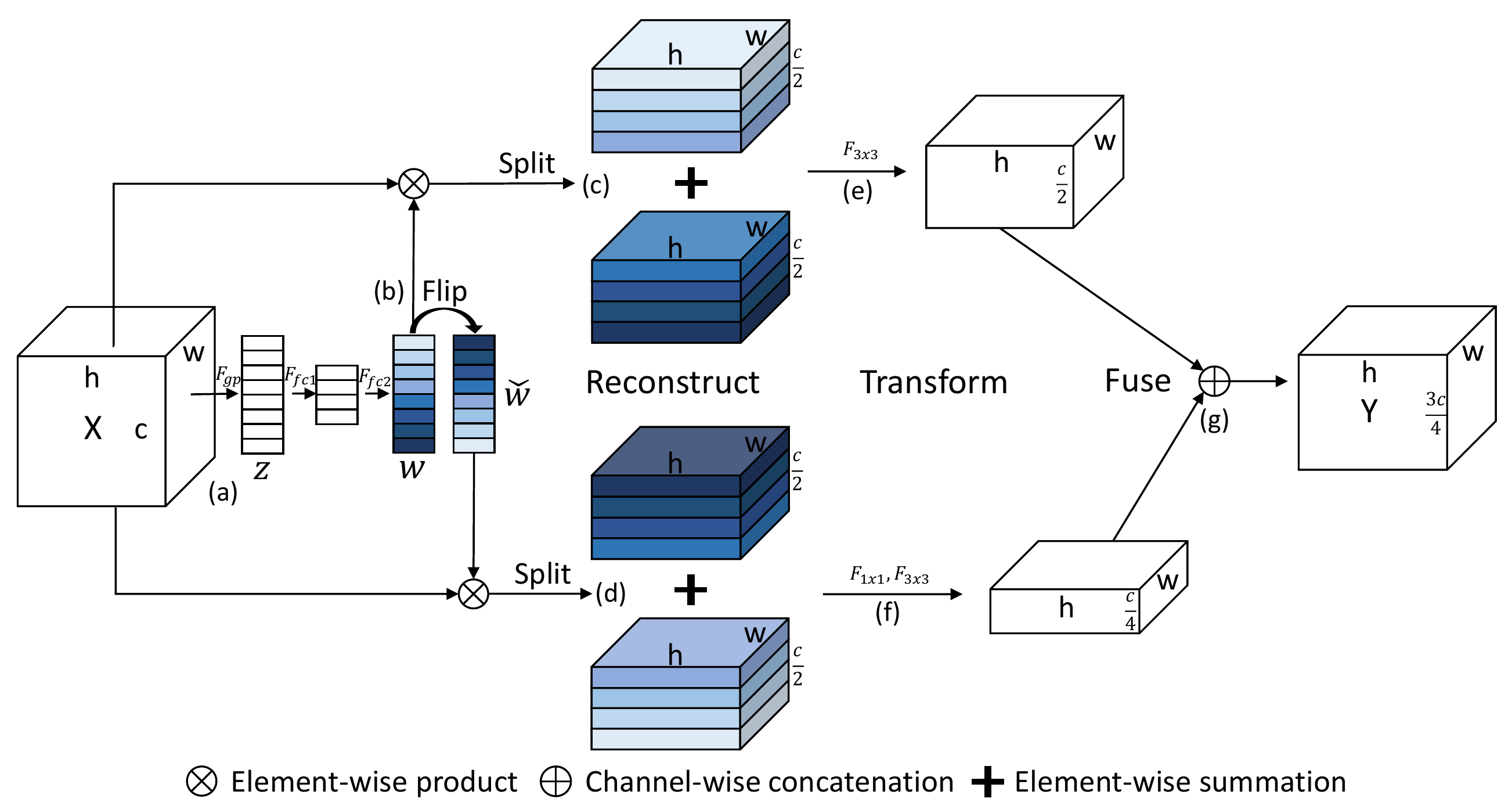}
\caption{The overview of our Slim Convolution. The module has two pathways. Each pathway does the weighted summation firstly and then transforms features. The features from two pathways are concatenated finally.}
\label{pip}
\end{figure}

The detailed procedure of SlimConv is depicted in Fig.~\ref{pip}, which mainly consists of three operations: \textbf{reconstruct}, \textbf{transform} and \textbf{fuse}. For reconstructing, we initially input a feedforward feature map into two pathways and modify a SE-module~\cite{hu2018squeeze} to get channel-wise weights of full channels (Fig.~\ref{pip} (a)). Secondly, for the top pathway, we multiply the input by the weights and halve the weighted features, followed by the element-wise summation of the pieces(Fig.~\ref{pip} (c)). For the lower pathway(Fig.~\ref{pip} (d)), we first flip the weights(Fig.~\ref{pip} (b)) and then conduct the same process as the top pathway. As the result, the feature channels are 2-fold after being reduced in each pathway. For transforming, a simple convolution layer with $3\times3$ kernel serves as the transformer for the top pathway(Fig.~\ref{pip} (e)). Meanwhile, a convolution layer with $1\times1$ kernel and a following one with $3\times3$ kernel are adopted for the bottom pathway(Fig.~\ref{pip} (f)). Here, the transformer with small kernel size also reduce channels by half. Finally, the features from two pathways are directly concatenated for the feature fusion(Fig.~\ref{pip} (g)). In this way, the SlimConv can reduce the number of channels substantially while retains the capability of representation according to our experiments. Notably, SlimConv is a plug-and-play module that can be applied to enhance the efficiency of various backbone architectures through simply replacing their original convolutions.

To evaluate the performance of SlimConv module, we conduct various experiments for common visual tasks on leading benchmarks, including ImageNet~\cite{russakovsky2015imagenet}, MS COCO2014~\cite{lin2014microsoft}, Pascal VOC2012 segmentation~\cite{hariharan2011semantic}, and Pascal VOC2007 detection~\cite{everingham2010pascal}. Experimental results show that the performance of our model is competitive comparing with most available state-of-the-art networks with respect to the accuracy and efficiency. Remarkably, the MobileNet(v2)~\cite{sandler2018mobilenetv2} equipped with SlimConv achieves $71.7\%$ top-1 accuracy with only $0.256$ GFLOPs and $3.38$M parameters, which shows $0.3\%$ higher performance with about $20\%$ lower cost and $4\%$ less parameters. Likewise, when equipping ShuffleNet(v2)~\cite{ma2018shufflenet}, the accuracy improves nearly $1\%$ while still reducing $40$ MFLOPs and $30K$ parameters. Meanwhile, we present ablation studies to demonstrate the effectiveness with respect to every part of our design. In addition, we theoretically discuss the compression of the features by analyzing the learnt weights which verifies its superiority.

In summary, our main contributions are as follows:

\begin{itemize}
  \item We design a plug-and-play module named SlimConv that can compress models and enhance representation ability of CNNs.
  \item We propose to reconstruct features to reduce the channel redundancy, within which a weights flipping operation can largely strengthen diversity of features.
  \item We integrate various CNN backbones with the proposed SlimConv. Meaningful improvements have been achieved through experiments on  challenging tasks, such as Image classification, Semantic segmentation and Object detection.
\end{itemize}

The rest of the paper is organized as follows: Sec.~\ref{Sec:relatedwork} introduces the related work in the area. Then Sec.~\ref{Sec:Method} explains the proposed SlimConv and its example bottleneck architecture. Experiments and analysis are presented in Sec.~\ref{Sec:Experiments}, and finally in Sec.~\ref{Sec:Conclusion}, concluding remarks are discussed.

\section{Related Work}\label{Sec:relatedwork}
\subsection{Efficient network architecture design}
Pioneering works on computer vision tasks achieved higher accuracy every year by prompting the network architecture to be deeper and more complex, such as AlexNet~\cite{krizhevsky2012imagenet} and VGG~\cite{simonyan2014very} on ImageNet competition. From hand-craft designs, the increase in the number of parameters and computational complexity made the improvement of accuracy less beneficial. InceptionNet~\cite{szegedy2015going} proposed Inception module to deepen the network with few added parameters. ResNet~\cite{he2016deep} and DenseNet~\cite{huang2017densely} utilized the efficient residual block by adopting shortcut connections. ResNeXt~\cite{xie2017aggregated} replaced traditional convolutions with group convolutions and introduced cardinality to increase model capacity. Res2Net~\cite{gao2019res2net} combined ResNet and ResNeXt and proposed an unusual multi-scale method. As the deployment for neural networks on terminal devices requires more lightweight models, the networks are encouraged to mobile-size such as SqueezeNet~\cite{iandola2016squeezenet}, ShuffleNet~\cite{ma2018shufflenet}, Xception~\cite{chollet2017xception} and MobileNet~\cite{sandler2018mobilenetv2}. Except for modifying backbones, some methods attempted to prune the trained models, such as \cite{han2016dsd, luo2018thinet, li2016pruning, han2015learning, he2017channel}, which pruned the inconsequential connections and weight to decrease the model size at a moderate accuracy loss. Recently, there has been a trend of neural architecture search in the research for designing more efficient CNN. These methods, such as NAS~\cite{zoph2018learning}, PNAS~\cite{liu2018progressive} and MNASNet\cite{tan2019mnasnet}, obtained the best network architecture by learning to explore the network structures, including width, depth, convolution kernels and connections. EfficientNet~\cite{tan2019efficientnet} and EffiecientDet~\cite{tan2019efficientdet} produced experimental evidence for scaling normal models to larger ones as the backbone with the method of NAS~\cite{zoph2018learning}, aiming at maximizing accuracy with limited resources. However, they need to mobilize large number of computing resources to automatically fulfill the task. Besides, the series of Slimmable Network~\cite{yu2018slimmable, yu2019universally, yu2019autoslim} was another kind of approach which proposed to learn a scale factor for each layer to reduce the network width while preserving the performance. It is so restrictive for the network architecture that it is difficult to search for ideal models for many tasks, and it needs iterative training procedures.

\subsection{Computational cost reduction}
Traditional convolution operation is created to extract local features as channel information and map their appearance to a feature map. When it comes to designing an effective lightweight CNN, efficient convolution operation is the most direct way. It can significantly reduce the channel redundancy in the network so that it does not need to prune or compress the model with extra computations after training. Furthermore, it can reorganize the features for performance improvements.  Based on the group convolutions~\cite{krizhevsky2012imagenet, xie2017aggregated}, ShuffleNet~\cite{ma2018shufflenet} proposed channel shuffle operation to build a desirable lightweight model, due to its improvement of the information flow across feature channels. MobileNet~\cite{sandler2018mobilenetv2} introduced depth-wise separable convolutions which demonstrated good representative ability than regular convolutions, thus to reduce the number of parameters and accelerate the training. OctConv~\cite{chen2019drop} and Multi-grid CNN~\cite{ke2017multigrid} respectively proposed octave convolution and multi-grid convolution to exploit multi-scale representations. Especially, octave convolution took more consideration regarding the efficient design to reduce the feature redundancy and strengthen the information exchange between channels between high or low frequencies. In contrast, our SlimConv can decrease the computational cost and storage at the same time by pruning a part of channels of a convolution layer.

\subsection{Attention mechanisms}
Attention has been widely used in many research fields such as salient object detection~\cite{fan2019shifting}, depth completion~\cite{qiu2019deeplidar}, image super-resolution\cite{zhang2018image} and facial expression recognition~\cite{marrero2019feratt}. Wang \etal~\cite{wang2017residual} proposed an attention module with Encoder-Decoder style for the image classification task. SENet~\cite{hu2018squeeze} introduced a lightweight attention module to re-calibrate the feature map by channel-wise weights. Besides channel-wise importance, CBAM~\cite{woo2018cbam} considered spatial attention and designed two sequential sub-modules including channel-wise and spatial attention. SKNet~\cite{li2019selective}, following InceptionNet and SENet, adopted soft-attention mechanism to make networks select the features with different receptive fields automatically. We seems like the first to introduce attention mechanism to the efficient module for channel reduction.

\section{Method}\label{Sec:Method}
\subsection{Slim Convolution}
Figure~\ref{pip} shows our pipeline of SlimConv, which includes two pathways and consists of three steps: \textit{Reconstruct}, \textit{Transform} and \textit{Fuse}. Note that, Fig.~\ref{pip} only shows our default setting, and it is flexible to expand to fit different bottlenecks.

\begin{figure*}[t]
\centering
\includegraphics[scale=0.45]{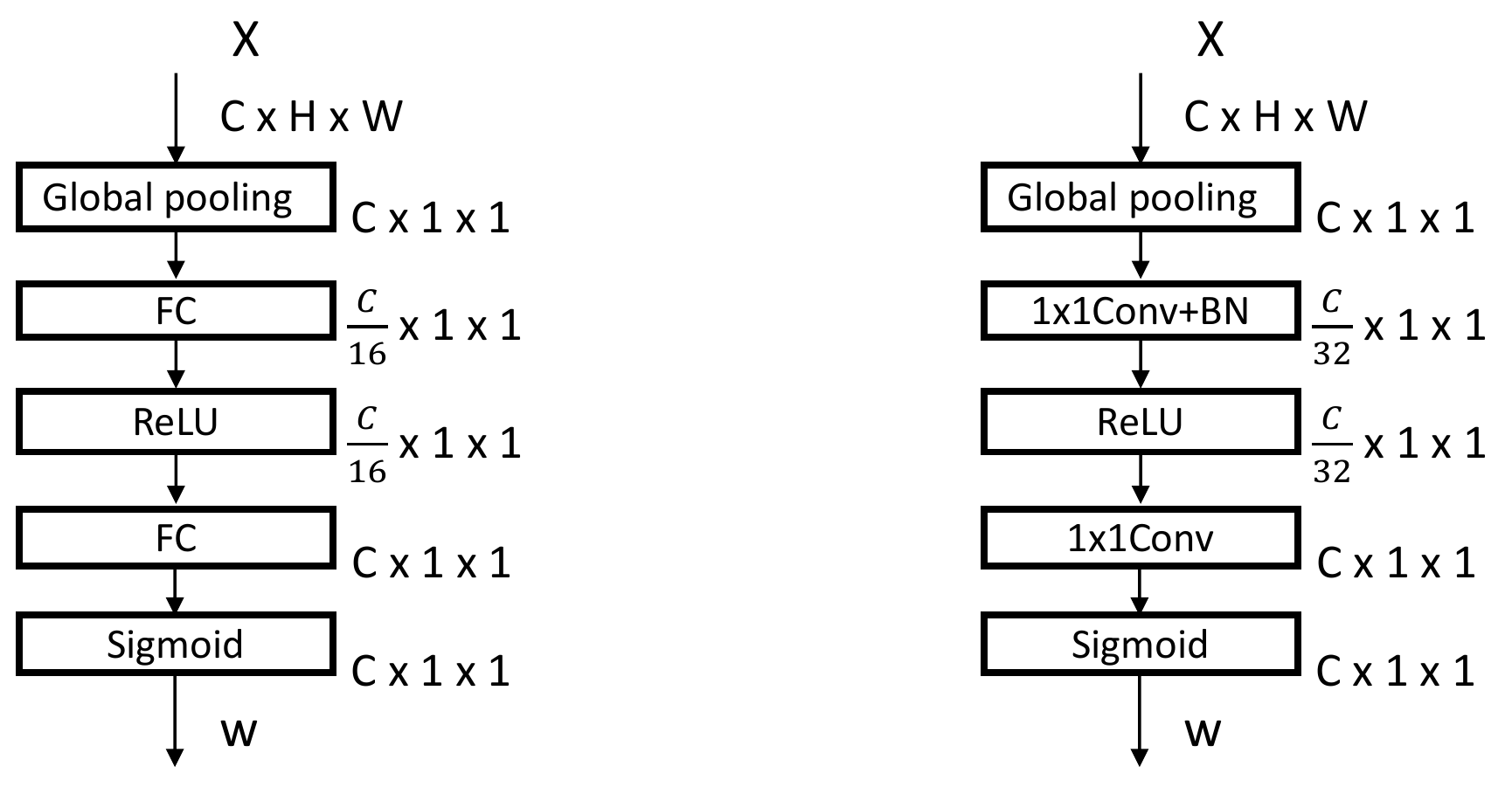}
\caption{\textbf{Modified SE-Module}. Left: the original SE-Module\cite{hu2018squeeze}. Right: we replace the FC layers with convolution layers and increase the reduction ratio to 32.}
\label{mse}
\end{figure*}

\subsubsection{Reconstruct.} Given an input feature map: $X \in \mathbb{R}^{C \times H \times W}$, we use a modified SE-module\cite{hu2018squeeze} to obtain channel-wise weights $w$. As illustrated in Fig. \ref{mse}, we replace the fully connected (fc) layer with the convolutional layer whose kernel size is $1\times1$, and use larger reduction ratio $32$ as the default setting. The whole process of acquiring $w$ can be expressed as:

\begin{equation}
    \begin{cases}
    z = F_{gp}(X) = \frac{1}{H \times W}\sum_{i=1}^H \sum_{j=1}^W X(i,j),
    \\
    w = \sigma(F_{fc2}(\delta(F_{fc1}(z)))).
    \end{cases}
\end{equation}
where $z \in \mathbb{R}^C$ contains channel-wise statistics, $\sigma$ refers to the sigmoid function and $\delta$ is the ReLU\cite{nair2010rectified} activation. $F_{fc1}$ and $F_{fc2}$ are convolution operations, $F_{fc1}$ includes the Batch Normalization\cite{ioffe2015batch}.

In the top pathway, we multiply features by $w$, yielding weighted features $X_w$. Then, we split $X_w$ into two parts ($X_w^1$, $X_w^2$), and sum them to compress the number of features to half:

\begin{equation}
    \begin{cases}
        X_w^{'} = X_w^1 + X_w^2, \\
        X_w = w * X, \\
        X_w^1 \cup X_w^2 = X_w.
    \end{cases}
\end{equation}

The compression can reduce redundant features, but may also result in the loss of valuable information. To deal with it, we propose the bottom pathway.

In the bottom pathway, we disrupt the order of feature weights through weights flipping. Further, we use flipped channel-wise weights $\check{w}$ to go through the same operations as the top pathway to obtain the half-channel features $X_{\check{w}}^{'}$.

\subsubsection{Transform.} We follow the bottleneck design rules of ResNet\cite{he2016deep} and conduct two transformers. The top transformer $F_{3 \times 3}$ is a convolution layer with kernel size 3. The bottom transformer contains two convolution layers $F_{1 \times 1},F_{3 \times 3}$ with kernel sizes 1 and 3 respectively. The convolution layer with kernel size 1 reduces the number of channels by half, and then followed by the convolution layer with kernel size 3.
\subsubsection{Fuse.}  We concatenate different features from two pathways to integrate information. Our SlimConv outputs features Y with channel number $\frac{3c}{4}$.

\subsection{Network Architecture}\label{NA}

\begin{figure*}[t]
\centering
\includegraphics[scale=0.4]{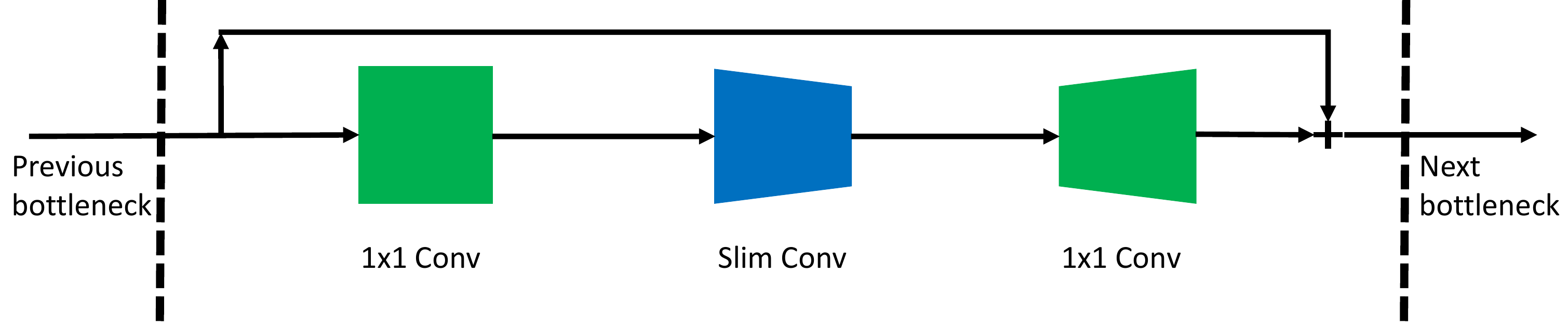}
\caption{\textbf{Modified Res-Bottleneck}.}
\label{mrb}
\end{figure*}

Since our proposed SlimConv takes $N$-channel features as input and output $\frac{N}{k}$-channel ($k \textgreater 1$, default is $\frac{4}{3}$) features, it has only one extra hyperparameter $k$ and can be easily integrated into many state-of-the-art CNN models, such as ResNet~\cite{he2016deep}, ResNeXt~\cite{xie2017aggregated}, DLA~\cite{yu2018deep}, MobileNet~\cite{sandler2018mobilenetv2}, ShuffleNet~\cite{ma2018shufflenet}, Big-Little Net~\cite{chen2018big}. We take ResNet~\cite{he2016deep} as an example.

As illustrated in Fig.~\ref{mrb}, the proposed SlimConv displaces the previous ordinary convolution layer which with kernel size 3 and decreases the number of output channels. So the input of the last convolution layer with kernel size 1 need to be changed accordingly.

Specially, when embedding SlimConv in DLA46-C~\cite{yu2018deep} and MobileNet V2~\cite{sandler2018mobilenetv2} which only have few channels in a convolution layer, we replace $\frac{C}{32}$ with $max(C/r,L)$ ($r$ and $L$ are fixed values) in our modified SE-Module.

\section{Experiments}\label{Sec:Experiments}
\subsection{Implementation Details}
Our proposed model and other state-of-the-art CNN-based models we used are all implemented by PyTorch\cite{paszke2019pytorch}. Similar to \cite{xie2017aggregated} , these models are trained on less than 8 GeForce RTX 2080 Ti GPUs. We mainly validate the effectiveness of our proposed model on $4$ challenging datasets:

\subsubsection{ImageNet.} We use the most popular dataset ImageNet~\cite{russakovsky2015imagenet} for all the experiments on the image classification. ImageNet is also a common benchmark, which contains 1.28 million images for training and 50k images for validation, all these images have labels from 1000 categories. We train the SlimConv-equipped models on training images, and pick the model with best top-1 error performance on validation images. We conduct the random-size cropping to 224 $\times$ 224 and random horizontal flipping~\cite{szegedy2015going}. For fair comparisons on all models, we use the same data argumentation and training strategy as~\cite{he2016deep}, \cite{sandler2018mobilenetv2} and \cite{chen2018big} respectively.

\subsubsection{Pascal VOC2012 Aug.} We evaluate the performance of our proposed model on semantic segmentation by using PASCAL VOC12 dataset~\cite{everingham2015pascal}. Following previous works, we use the augmented version of PASCAL VOC12 dataset~\cite{hariharan2011semantic} which contains $10,582$ training images and $1,449$ validating images from $21$ classes. We use state-of-the-art method Deeplab v3+\cite{chen2018encoder} as the segmentation framework and the same implementation details for all models.

\subsubsection{MS COCO2014 \& PASCAL VOC2007.} For object detection, we evaluate our SlimConv on MS COCO dataset~\cite{lin2014microsoft} and PASCAL VOC2007 dataset~\cite{everingham2010pascal}. We take the widely used method Faster RCNN~\cite{ren2015faster} as the detection framework and use the same strategy to train and test models.

\subsection{Image Classification}
We perform image classification experiments on the ImageNet dataset~\cite{russakovsky2015imagenet} to evaluate our module. Our SlimConv is a plug-and-play module, we embed it into public competitive models for comparison.

\begin{table}[t]
\begin{center}
\begin{tabular}{l|c|c|c}
\hline
Model & Top-1 Error & FLOPs($10^9$) & Params($10^6$) \\
\hline
\hline
ResNet-50 & 23.85 & 4.12 & 25.56\\
AutoSlim-ResNet-50 & 24.00 & 3.00 & 23.10\\
SE-ResNet-50 & \textbf{23.29} & 4.12 & 28.09\\
Sc-ResNet-50(ours) & \textbf{23.29} & \textbf{2.69} & \textbf{16.76}\\
\hline
ResNet-101 & 22.63 & 7.84 & 44.55\\
DenseNet-161 & 22.35 & 7.82 & 28.68\\
SE-ResNet-101 & 22.38 & 7.85 & 49.33\\
Sc-ResNet-101(ours) & \textbf{22.16} & \textbf{4.87} & \textbf{27.96}\\
\hline
ResNeXt-50(32$\times$4d) & 22.38 & 4.27 & 25.03\\
Sc-ResNeXt-50(32$\times$4d)(ours) & \textbf{22.03} & \textbf{3.86} & \textbf{22.49}\\
\hline
ResNeXt-101(32$\times$4d) & 21.20 & 8.03 & 44.18\\
Sc-ResNeXt-101(32$\times$3d,k=2)(ours) & \textbf{21.18} & \textbf{4.64} & \textbf{23.70}\\
\hline
\end{tabular}
\end{center}
\caption{Performance comparison for ResNet~\cite{he2016deep}, ResNeXt~\cite{xie2017aggregated}, DenseNet~\cite{huang2017densely}, SE-ResNet~\cite{hu2018squeeze}, neural-architecture-search method AutoSlim~\cite{yu2019autoslim} and our integrated models on ImageNet.}
\label{cp1}
\end{table}

\subsubsection{Comparing with middle sized models.} Table~\ref{cp1} reports $4$ group of results according to the complexity. Models that equipped with our SlimConv contains the prefix `Sc' in all the tables. In the first group, our integrated Sc-ResNet-50 achieves almost $0.6\%$ better accuracy, $35\%$ less FLOPs and parameters than non-equipped original ResNet-50~\cite{he2016deep}. With the modified SE-Mouble, our Sc-ResNet-50 shows the same accuracy performance as SE-ResNet-50~\cite{hu2018squeeze}, but cost over $40\%$ less parameters. When compared to the SOTA neural-architecture-search model AutoSlim\cite{yu2019autoslim}, our Sc-ResNet-50 also achieves $0.7\%$ better accuracy, $10.3\%$ less FLOPs and $27.4\%$ less parameters. In the second group, we consider deeper models and take ResNet-101~\cite{he2016deep} as the basic model. Our integrated Sc-ResNet-101 also achieves nearly $0.5\%$ less top-1 error than the basic model, while reducing FLOPs and parameters simultaneously by almost $38\%$. Our Sc-ResNet-101 is more efficient than DenseNet-161~\cite{huang2017densely} and SE-ResNet-101\cite{hu2018squeeze}. In the third group, we add our module to ResNeXt-50~\cite{xie2017aggregated}. We do the same group operation in the last two convolution layers with kernel size 3$\times$3. Due to our SlimConv can reduce channels, the width of each group is also decreased. Our integrated Sc-ResNeXt-50 achieves almost $0.4\%$ better accuracy while also reduces the computational cost and storage by $10\%$. In the last group, we change our hyperparameter $k$ to $2$, reducing the width from $128$ to $96$ during the integration to ResNeXt-101\cite{xie2017aggregated}. Even though having a thinner architecture, our Sc-ResNeXt-101 achieves a slightly improved accuracy, over $42\%$ less FLOPs and $46.4\%$ less parameters than the wider basic model.

\begin{figure}[t]
	\centering
	\begin{tabular}{*{8}{c@{\hspace{1px}}}}
		\raisebox{0.0\height}{\rotatebox{90}{\small{~ResNet-50}}}&
		\includegraphics[width=0.13\textwidth]{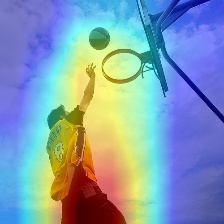}&
		\includegraphics[width=0.13\textwidth]{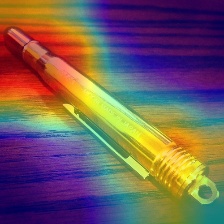}&
		\includegraphics[width=0.13\textwidth]{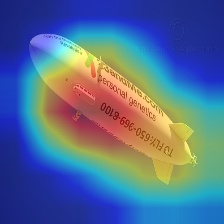}&
		\includegraphics[width=0.13\textwidth]{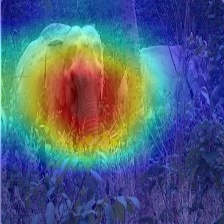}&
        \includegraphics[width=0.13\textwidth]{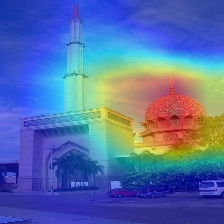}&
		\includegraphics[width=0.13\textwidth]{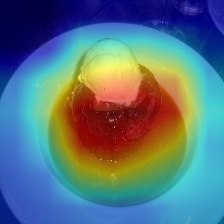}&
		\includegraphics[width=0.13\textwidth]{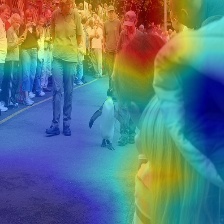}
		\\
		\vspace{-1.5pt}
        \raisebox{0.0\height}{\rotatebox{90}{\small{Sc-ResNet-50}}}&
		\includegraphics[width=0.13\textwidth]{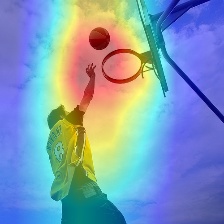}&
		\includegraphics[width=0.13\textwidth]{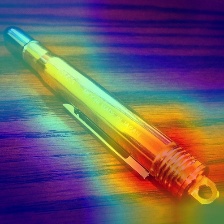}&
		\includegraphics[width=0.13\textwidth]{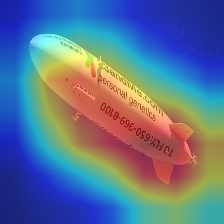}&
		\includegraphics[width=0.13\textwidth]{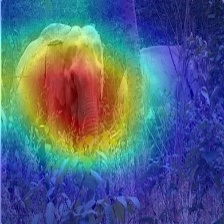}&
        \includegraphics[width=0.13\textwidth]{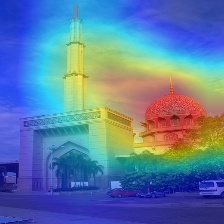}&
		\includegraphics[width=0.13\textwidth]{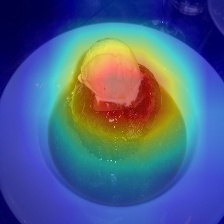}&
		\includegraphics[width=0.13\textwidth]{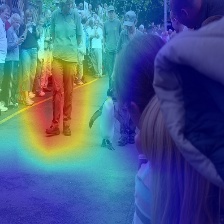}
		\\
		\vspace{-1.5pt}
		\raisebox{0.5\height}{\small{ }} &
		\raisebox{0.5\height}{\small{~Basketball}} &
		\raisebox{0.5\height}{\small{~Ballpoint}} &
		\raisebox{0.5\height}{\small{~Airship}} &
		\raisebox{0.5\height}{\small{~Elephant}} &
		\raisebox{0.5\height}{\small{~Mosque}} &
		\raisebox{0.5\height}{\small{~Ice cream}} &
		\raisebox{0.5\height}{\small{~Penguin}}
	\end{tabular}
	\caption{Grad-CAM\cite{selvaraju2017grad} visual comparison for ResNet-50\cite{he2016deep} and our Sc-ResNet-50 on ImageNet.}
	\label{gc}
\end{figure}

\subsubsection{Comparing with multi-scale models.} Multi-scale strategy~\cite{chen2018big,chen2019drop} is effective for image classification,
Table~\ref{cp2} reports the results. Here, we chose a model from Big-Little Net~\cite{chen2018big}, named as bL-ResNet-50, to be our basic model, where `bL' stands for Big-little. As seen, SlimConv equipped model, Sc-bL-ResNet-50, achieves nearly $0.4\%$ better accuracy, $23.2\%$ less FLOPs and over one third less parameters than the basic model. Compared with the SOTA model Oct-ResNet-50, our model also achieves almost $0.3\%$ less top-1 error, less FLOPs and over $30\%$ less parameters.

\begin{table}[t]
\begin{center}
\begin{tabular}{l|c|c|c}
\hline
Model & Top-1 Error & FLOPs($10^9$) & Params($10^6$) \\
\hline
\hline
ResNet-50 & 23.85 & 4.12 & 25.56\\
\hline
\hline
bL-ResNet-50($\alpha = 2,\beta = 4$) & 22.69 & 2.85 & 26.69 \\
Oct-ResNet-50($\alpha=0.5$) & 22.60 & 2.40 & 25.60 \\
Sc-bL-ResNet-50($\alpha = 2,\beta = 4$)(ours) & \textbf{22.34} & \textbf{2.19} & \textbf{17.77} \\
\hline
\end{tabular}
\end{center}
\caption{Performance comparison for multi-scale models: bL-ResNet-50~\cite{chen2018big}, Oct-ResNet-50~\cite{chen2019drop} and our Sc-bL-ResNet-50 on ImageNet.}
\label{cp2}
\end{table}

\begin{table}[t]
\begin{center}
\begin{tabular}{l|c|c|c}
\hline
Model & Top-1 Error & FLOPs($10^8$) & Params($10^6$) \\
\hline
\hline
Squeezenet-B & 39.60 & 7.20 & 1.20 \\
DLA-46-C & 35.96 & 5.90 & 1.31 \\
Sc-DLA-46-C(r=8, L=16)(ours) & \textbf{35.69} & \textbf{4.98} & \textbf{0.97} \\
\hline
\hline
1.0 ShuffleNet(v2) & 32.01 & 1.51 & 2.28 \\
1.0 Sc-ShuffleNet(v2)(ours) & \textbf{31.08} & \textbf{1.47} & \textbf{2.25} \\
\hline
1.0 MobileNet(v2) & 28.54 & 3.20 & 3.51\\
1.0 Sc-MobileNet(v2, k=$\frac{8}{3}$, r=24, L=6)(ours) & \textbf{28.26} & \textbf{2.56} & \textbf{3.38} \\
\hline
\end{tabular}
\end{center}
\caption{Performance comparison for lightweight models(Squeezenet~\cite{iandola2016squeezenet}, DLA~\cite{yu2018deep}, MobileNet v2~\cite{sandler2018mobilenetv2} and ShuffleNet v2~\cite{ma2018shufflenet}) and our integrated models on ImageNet.}
\label{cp3}
\end{table}

\subsubsection{Comparing with lightweight models.} We conduct 3 groups of experiments to test our performances on lightweight models. Table~\ref{cp3} reports the results. In the first group, we choose DLA-46-C\cite{yu2018deep} as the baseline and implement an efficient model with parameters less than 1 MB. Our integrated Sc-DLA-46-C achieves almost $0.3\%$ better accuracy and $15.6\%$ less FLOPs with only $0.97$ MB parameters than the baseline. When compared to Squeezenet~\cite{hu2018squeeze}, the SlimConv embedded model has nearly $4\%$ improvement in terms of accuracy, $30.8\%$ less FLOPs and $19.2\%$ less parameters than it. We also choose the most popular lightweight models~\cite{ma2018shufflenet,sandler2018mobilenetv2} as the baseline models. In the second group, our integrated Sc-ShuffleNet performs nearly $1\%$ better accuracy, less computational cost and parameters than ShuffleNet~\cite{ma2018shufflenet}. In the third group, our Sc-MobileNet also achieves almost $0.3\%$ less top-1 error, $20\%$ less FLOPs and less parameters than the basic MobileNet~\cite{sandler2018mobilenetv2}.

\begin{figure}[t]
	\centering
	\begin{tabular}{*{4}{c@{\hspace{0px}}}}
		\raisebox{0.5\height}{\small{~Input}} &
		\raisebox{0.5\height}{\small{~GT}} &
		\raisebox{0.5\height}{\small{~MobileNet v2}} &
		\raisebox{0.5\height}{\small{~Sc-MobileNet v2}}
		\\
		\vspace{-1.5pt}
		\includegraphics[width=0.23\textwidth]{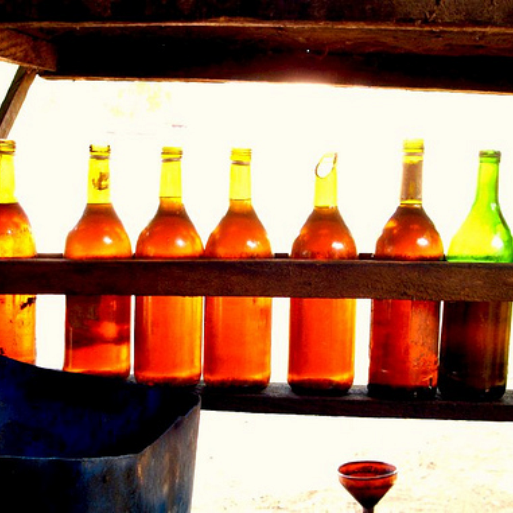}&
		\includegraphics[width=0.23\textwidth]{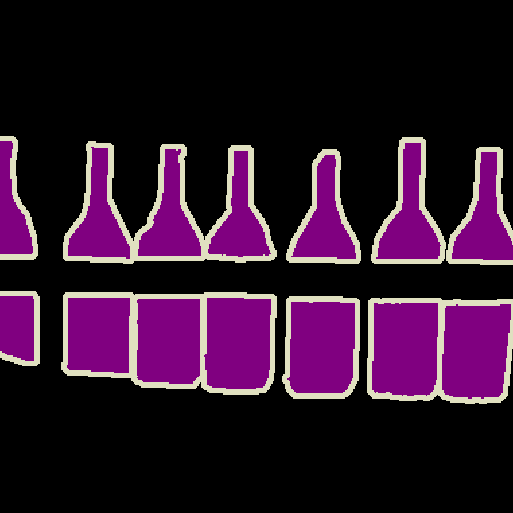}&
		\includegraphics[width=0.23\textwidth]{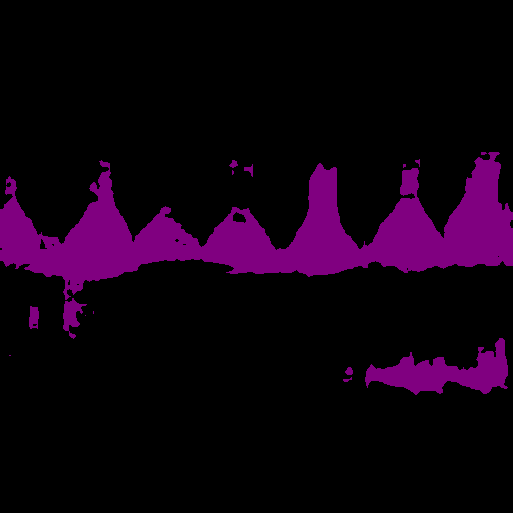}&
		\includegraphics[width=0.23\textwidth]{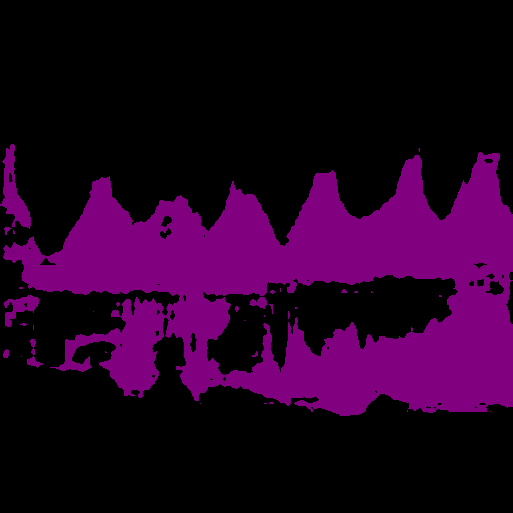}
		\\
		\vspace{-1.5pt}
		\includegraphics[width=0.23\textwidth]{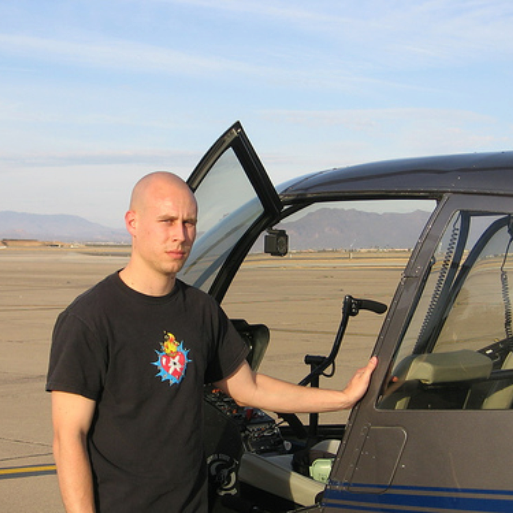}&
		\includegraphics[width=0.23\textwidth]{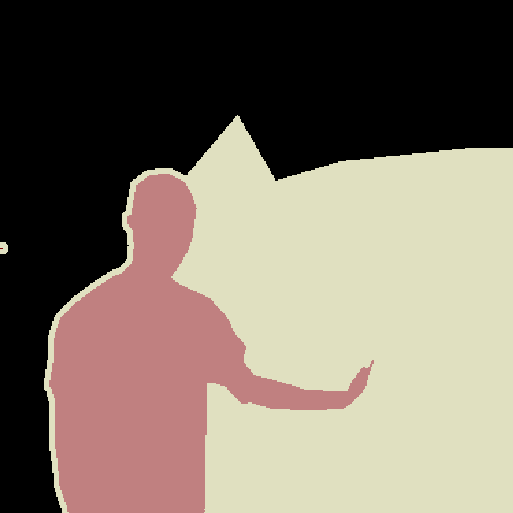}&
		\includegraphics[width=0.23\textwidth]{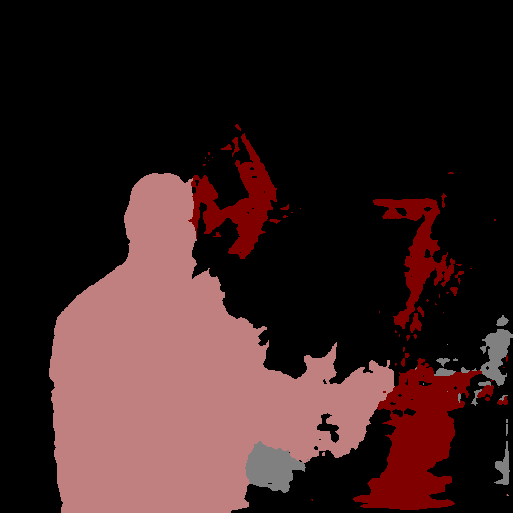}&
		\includegraphics[width=0.23\textwidth]{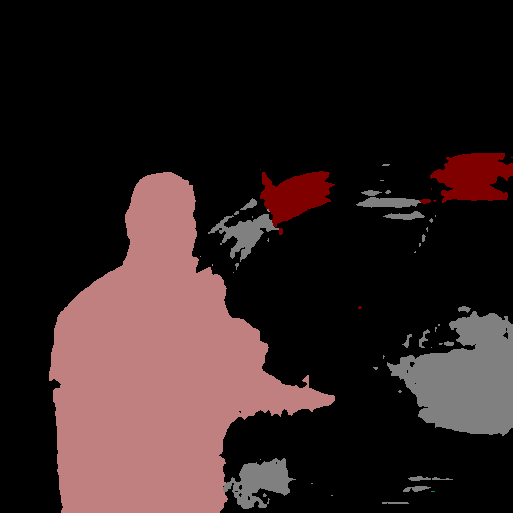}
		\\
		\raisebox{0.5\height}{\small{~Input}} &
		\raisebox{0.5\height}{\small{~GT}} &
		\raisebox{0.5\height}{\small{~ResNet-101}} &
		\raisebox{0.5\height}{\small{~Sc-ResNet-101}}
		\\
		\vspace{-1.5pt}
		\includegraphics[width=0.23\textwidth]{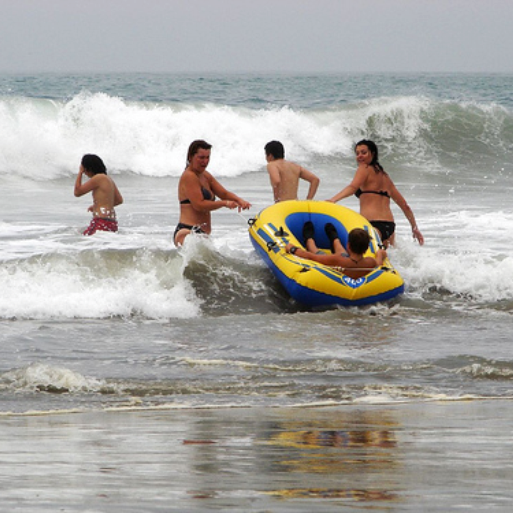}&
		\includegraphics[width=0.23\textwidth]{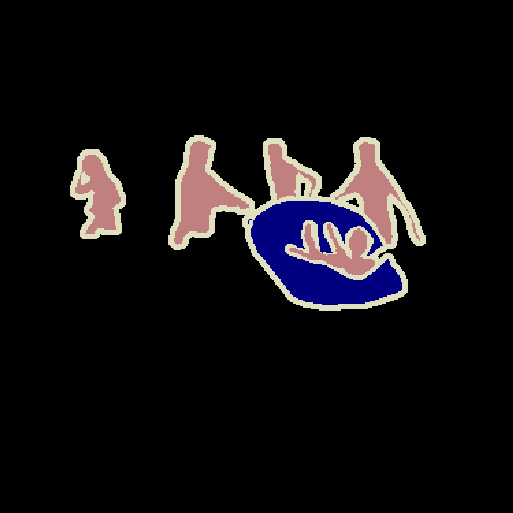}&
		\includegraphics[width=0.23\textwidth]{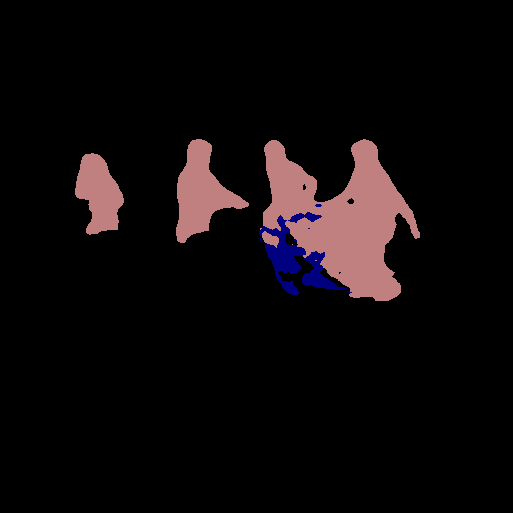}&
		\includegraphics[width=0.23\textwidth]{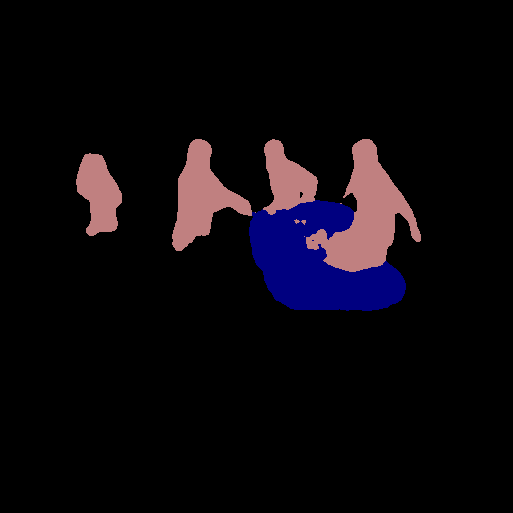}
		\\
		\vspace{-1.5pt}
		\includegraphics[width=0.23\textwidth]{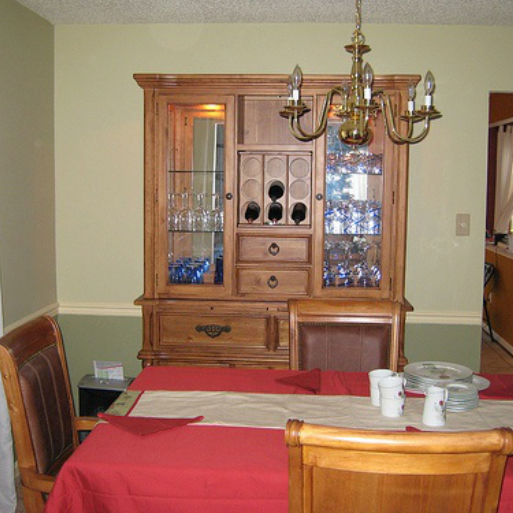}&
		\includegraphics[width=0.23\textwidth]{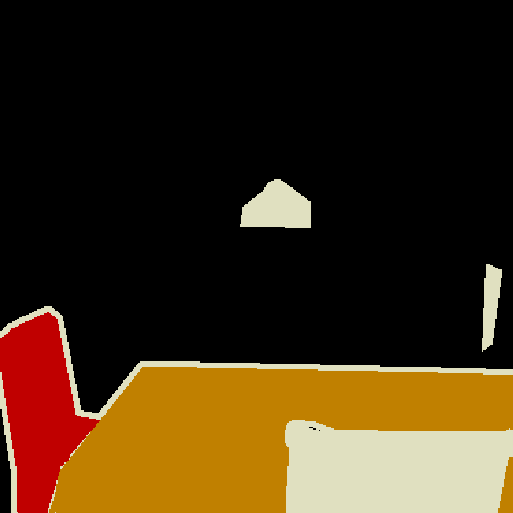}&
		\includegraphics[width=0.23\textwidth]{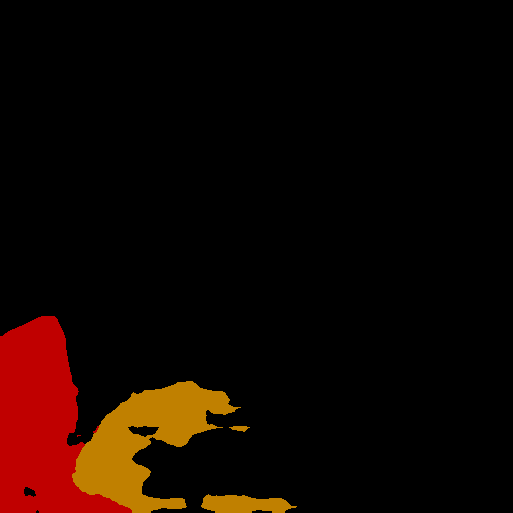}&
		\includegraphics[width=0.23\textwidth]{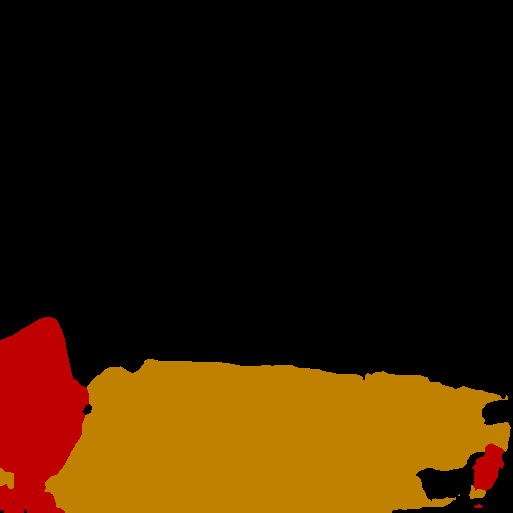}
	\end{tabular}
	\caption{Visual comparison on semantic segmentation for MobileNet v2\cite{sandler2018mobilenetv2}, ResNet\cite{he2016deep} and our integrated models on Pascal VOC2012.}
	\label{ss}
\end{figure}

\subsubsection{Class activation mapping.} We present some results of class activation mapping by Grad-CAM~\cite{selvaraju2017grad} to visualize where CNNs pay more attention for image classification. The visualization examples are shown in Fig.~\ref{gc}, the lighter the area, the more attention the network has. When compared to the results of ResNet-50~\cite{he2016deep}, the results of our integrated Sc-ResNet-50 concentrates on small objects such as `Basketball' and `Ice cream'. Even in complex scenes, the SlimConv equipped models can still concentrate on the area close to the object while ResNet-50 has been distracted. For large objects such as `Ballpoint', `Airship', `Elephant' and `Mosque', our activation maps are more accurate, which coverage of objectives comprehensively than the basic results.

\subsection{Semantic Segmentation}
Figure~\ref{gc} also shows that our module can strengthen networks for precisely localizing the region of objects. This characteristic makes our SlimConv has potential to improve the performance of models in the semantic segmentation task.

\begin{table}[h]
\begin{center}
\begin{tabular}{l|c|c|c}
\hline
Backbone & Mean IoU ($\%$) & FLOPs($10^9$) & Params($10^6$) \\
\hline
\hline
1.0 MobileNet(v2) & \textbf{69.3} & 14.20 & 5.23\\
1.0 Sc-MoblieNet(v2)(ours) & 69.2 & \textbf{13.80} & \textbf{5.10} \\
\hline
\hline
ResNet-50 & 76.4 & 62.60 & 39.76\\
Sc-ResNet-50(ours) & \textbf{76.8} & \textbf{50.18} & \textbf{30.96}\\
\hline
ResNet-101 & 77.8 & 83.31 & 58.75\\
Sc-ResNet-101(ours) & \textbf{78.5} & \textbf{62.32} & \textbf{42.16}\\
\hline
\end{tabular}
\end{center}
\caption{Performance comparison on semantic segmentation for MobileNet v2\cite{sandler2018mobilenetv2}, ResNet\cite{he2016deep} and our integrated models on Pascal VOC2012.}
\label{sst}
\vspace{-2.0em}
\end{table}

We replace the backbone network of Deeplab v3+~\cite{chen2018encoder} with MobileNet~\cite{sandler2018mobilenetv2}, ResNet~\cite{he2016deep} and our integrated models, and conduct three groups of comparisons. The results are shown in Table~\ref{sst}. In the group 1, our Sc-MobileNet achieves slightly worse mean IoU but less FLOPs and parameters than MobileNet~\cite{sandler2018mobilenetv2}.
As shown in Fig.~\ref{ss}, the SlimConv equipped model tends to correctly handle occluded scenes.
In the group 2, our Sc-ResNet-50 outperforms its baseline by $0.4\%$ on mean IoU, and also achieves almost $20\%$ less FLOPs and over $22\%$ less parameters than ResNet-50. In the last group, our integrated Sc-ResNet-101 achieves $0.7\%$ better mean IoU, $25.2\%$ less computational cost and $28.2\%$ less storage than the reference ResNet-101. As illustrated in Fig.~\ref{ss}, our Sc-ResNet-101 has the closest results to ground-truth while the coverage of objects is either more or less for ResNet-101 based results.

\subsection{Object Detection}
\vspace{-.5em}
We evaluate our module on two kinds of dataset for the object detection task. The results are shown in Table~\ref{od} and Table~\ref{od2}. We adopt backbone network of the widely used ResNet-101~\cite{he2016deep} v.s. our Sc-ResNet-101. Our SlimConv based model achieves $0.2\%$ better average precision than the original ResNet-101, reducing FLOPs by $34.1\%$ and parameters by $35\%$ on the PASCAL VOC2007 dataset~\cite{everingham2010pascal}. On the other popular dataset COCO2014~\cite{lin2014microsoft}, our Sc-ResNet-101 achieves $1.4\%$ better average precision with over $52$GB computational cost decreased, and $34.5\%$ less parameters than ResNet-101.
\vspace{-1.5em}
\begin{table}[h]
\begin{center}
\begin{tabular}{l|c|c|c}
\hline
Backbone & mAP & FLOPs($10^9$) & Params($10^6$) \\
\hline
\hline
ResNet-101 & 75.0 & 148.93 & 45.26\\
Sc-ResNet-101(ours) & \textbf{75.2} & \textbf{98.16} & \textbf{29.44}\\
\hline
\end{tabular}
\end{center}
\caption{Detection performance on PASCAL VOC2007 dataset~\cite{everingham2010pascal}. Note that FLOPs and Params are calculated when the size of input image is $850\times600$.}
\label{od}
\vspace{-5.0em}
\end{table}

\begin{table}[h]
\begin{center}
\begin{tabular}{l|c|c|c}
\hline
Backbone & AP@(IoU=0.50:0.95) & FLOPs($10^9$) & Params($10^6$) \\
\hline
\hline
ResNet-101 & 33.9 & 153.59 & 45.86\\
Sc-ResNet-101(ours) & \textbf{35.3} & \textbf{101.36} & \textbf{30.04}\\
\hline
\end{tabular}
\end{center}
\caption{Detection performance on COCO2014 minival dataset~\cite{lin2014microsoft}. Note that FLOPs and Params are calculated when the size of input image is $600\times899$.}
\label{od2}
\vspace{-5.0em}
\end{table}

\subsection{Ablation Studies}
\vspace{-2.0em}
\begin{table}[H]
\begin{center}
\begin{tabular}{l|c|c|c}
\hline
Model & Top-1 Error & FLOPs($10^9$) & Params($10^6$) \\
\hline
\hline
ResNet-50 & 23.85 & 4.12 & 25.56\\
ResNet-50@256 & 23.28 & 5.38 & 25.56\\
\hline
\hline
Sc-ResNet-50(w/o flipping) & 23.71 & 2.69 & 16.76 \\
Sc-ResNet-50(only filpping)  & 23.93 & 2.69 & 16.76 \\
Sc-ResNet-50 & 23.29 & 2.69 & 16.76 \\
Sc-ResNet-50(cosine)  & 22.77 & 2.69 & 16.76 \\
Sc-ResNet-50@256 & \textbf{22.52} & 3.51 & 16.76 \\
\hline
\end{tabular}
\end{center}
\caption{Performance comparison for our Sc-ResNet-50 with different settings on ImageNet dataset. For @256, we directly change the size of input image from $224\times224$ to $256\times256$ on ImageNet.}
\label{ab}
\end{table}

To explore the effectiveness of our different design choice, we conduct ablation studies on the ImageNet dataset~\cite{russakovsky2015imagenet} with ResNet-50~\cite{he2016deep} as the baseline. The experimental results are shown in Table~\ref{ab}.

Firstly, we drop the flipping operation and make the bottom pathway use the same weights as the top pathway. We find that Sc-ResNet-50 without flipping can still achieve $0.14\%$ better accuracy than the baseline. It shows that our method possess strong robustness. Secondly, we replace the learned weights $w$ with the flipped weights $\check{w}$ so that all weights used are flipped. In this case, the SlimConv with only flipped weights makes a little worse accuracy than the baseline. The first case converges faster but also over-fits earlier than the second one during training. When compared to our proposed model, which outperforms other settings because of the weights flipping. Furthermore, we try to use the cosine learning rate. Our Sc-ResNet-50(cosine) achieves better results than before. Last but not least, we increase the size of input image to $256\times256$ and test the performance of the pre-trained models. Our Sc-ResNet-50@256 achieves $0.76\%$ better accuracy and almost $35\%$ less computational cost than ResNet-50@256 as illustrated in Table~\ref{ab}.

\subsection{Discussion}
\begin{figure}[t]
	\centering
	\begin{tabular}{*{3}{c@{\hspace{1px}}}}
		\includegraphics[width=0.45\textwidth]{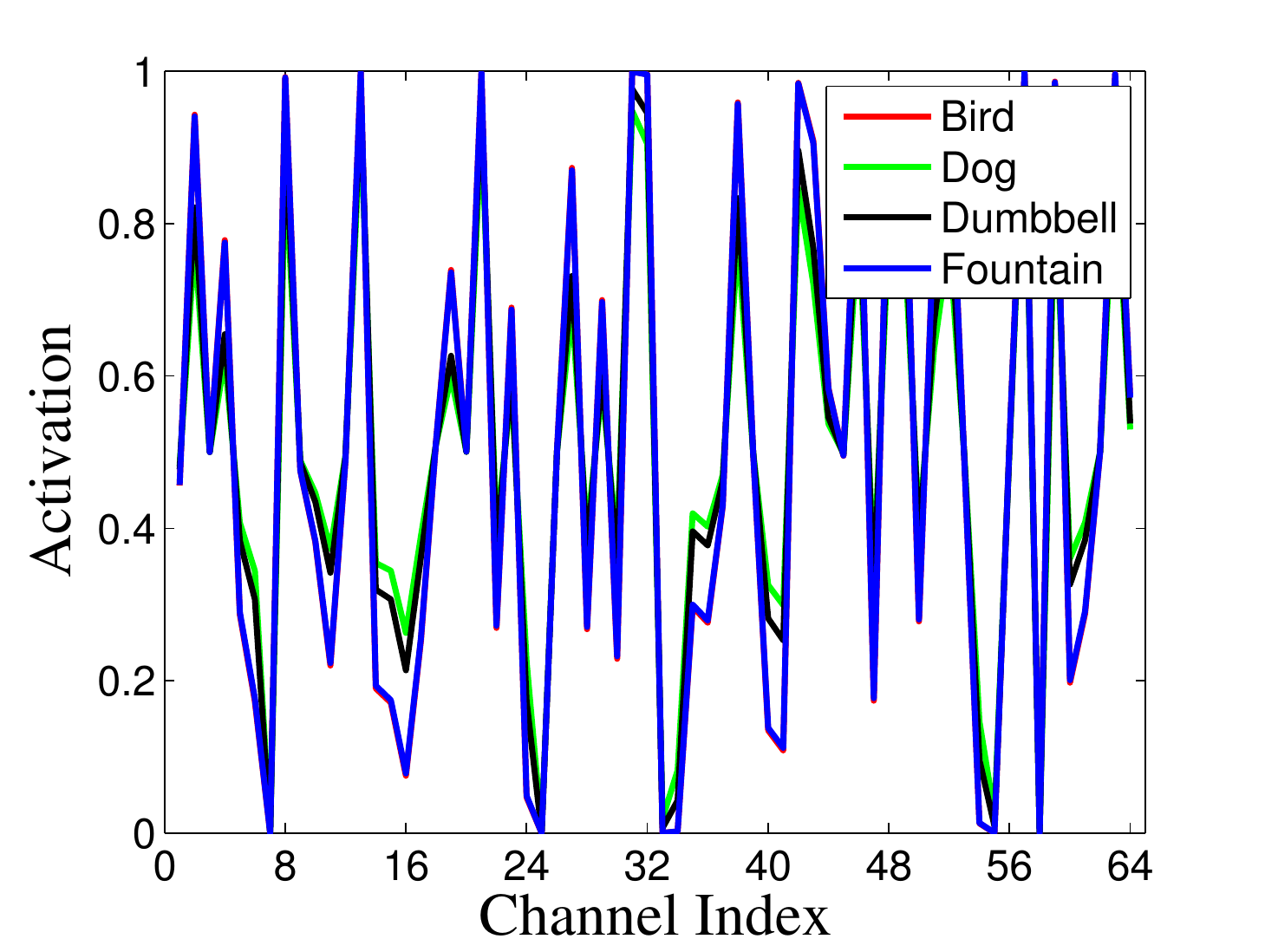}
		&
		\includegraphics[width=0.45\textwidth]{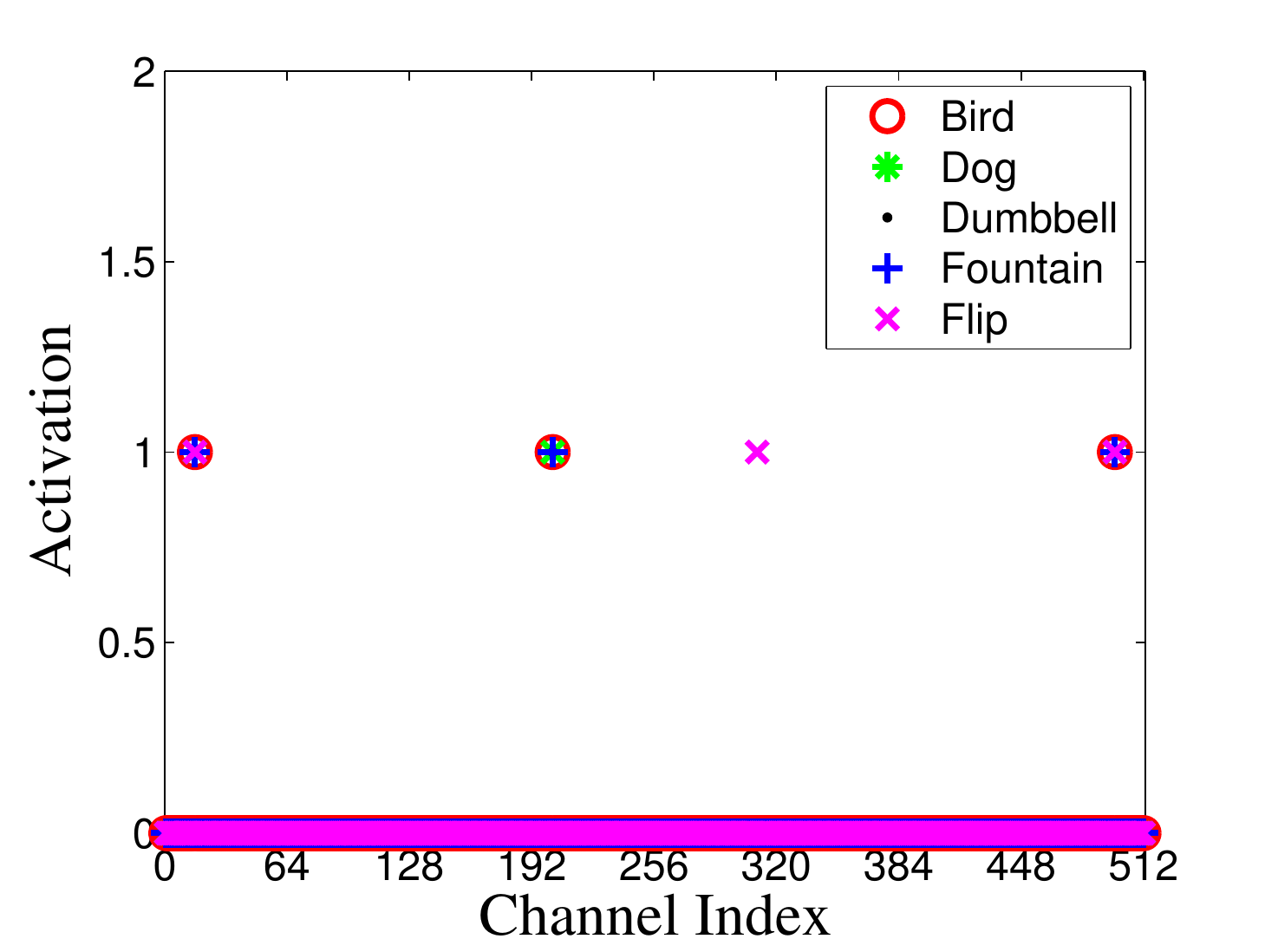}
		\\
		\vspace{-1.5pt}
		\raisebox{0.5\height}{\small{~(a) Sc\_2\_1}}
		&
		\raisebox{0.5\height}{\small{~(b) Sc\_5\_3}}
	\end{tabular}
	\caption{Last activations(w) of the modified SE-module in our Sc-ResNet-50 on ImageNet dataset. Two set of activations are named by following the scheme: \textit{Sc\_stageID\_blockID}. The indexes with weight one are: 16, 203, 310 and 497. `Flip' means the flipped weights.}
	\label{sco}
\end{figure}

We analyze the learned weights by inputting different objects to investigate how information of features compressed. The results are shown in the Fig.~\ref{sco}. At the first block of the second stage, the values of output weights are varied from zero to one. But at the last block named Sc\_5\_3, the weights just have zero and one. Interestingly, the weights of different objects all have the same values. This finding indicates that different features in the channel can be integrated efficiently by our proposed SlimConv.
In addition, the learned weights of our SlimConv module becomes extremely sparse at the high-level blocks, such as Sc\_5\_3. Then, the features with weight zero can be abandoned to save more parameters and FLOPs. As illustrated in Fig.~\ref{pip}, if we drop the input features of weight zero, input features of the top convolutional layer (Fig.~\ref{pip}~(e)) and the first bottom convolutional layer (Fig.~\ref{pip}~(f)) will both be decreased. Theoretically, some parameters and FLOPs can still be reduced.

\begin{figure}[t]
	\centering
	\begin{tabular}{*{3}{c@{\hspace{0px}}}}
		\includegraphics[width=0.33\textwidth]{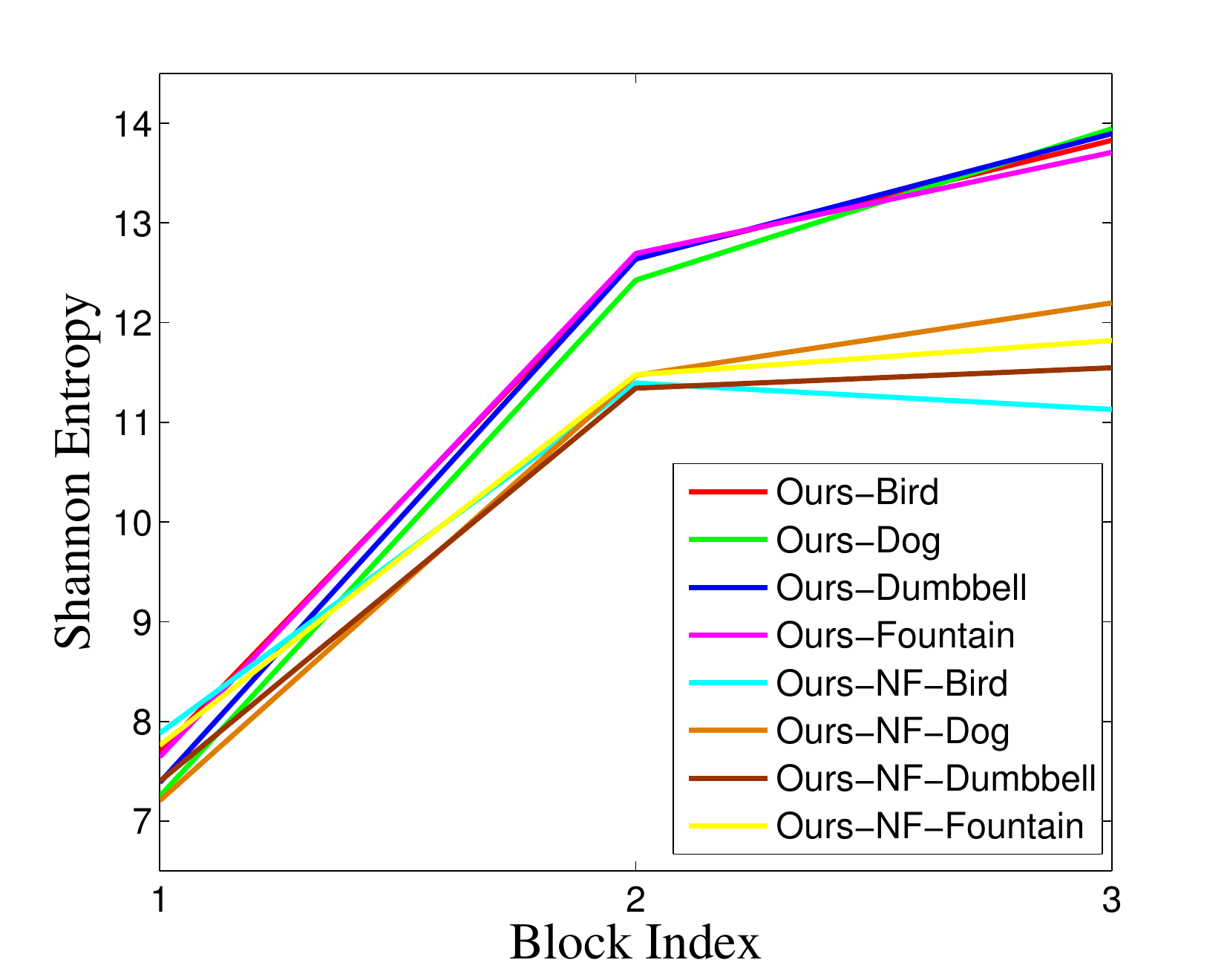}
		&
		\includegraphics[width=0.33\textwidth]{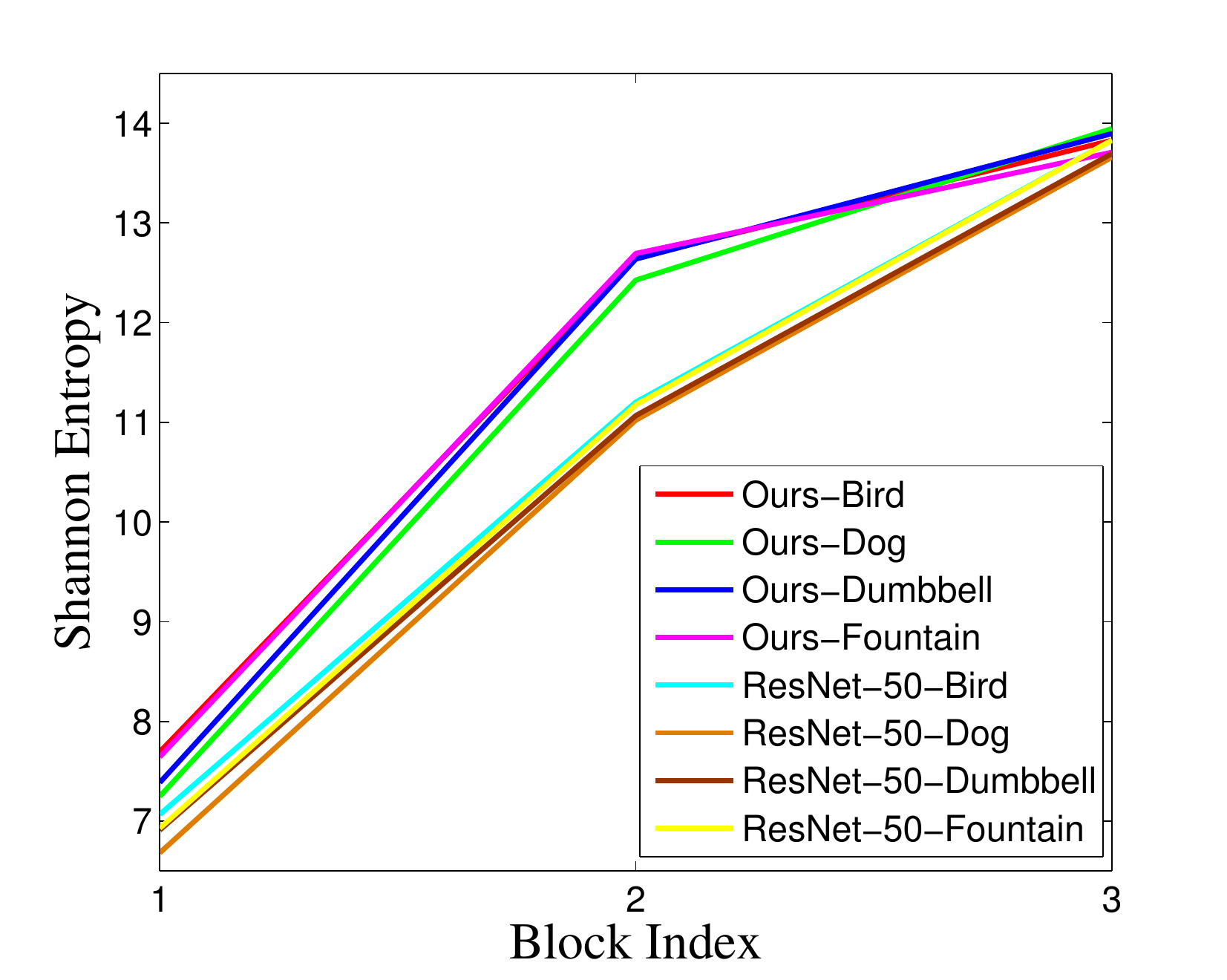}
		&
		\includegraphics[width=0.33\textwidth]{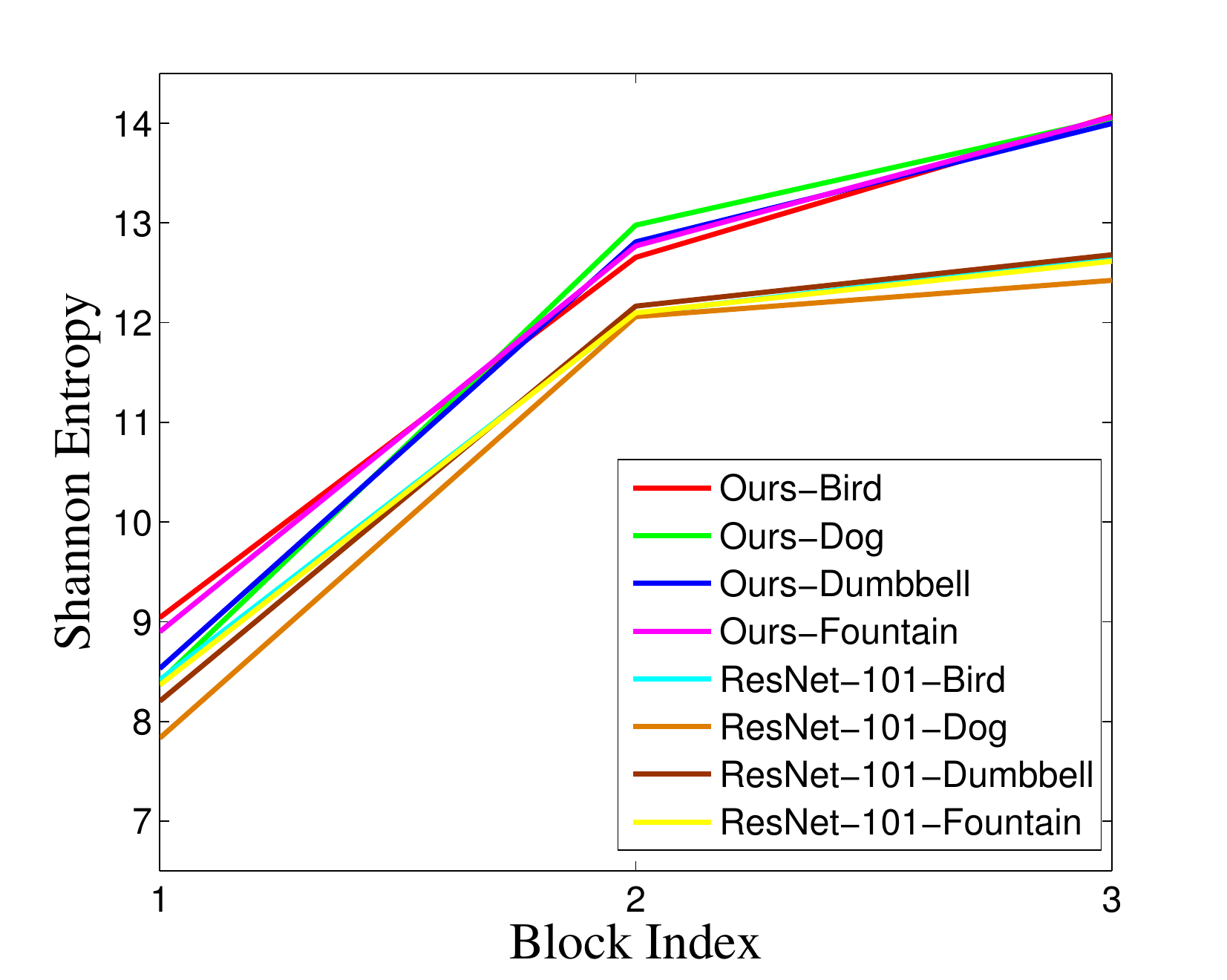}
		\\
		\vspace{-1.5pt}
		\raisebox{0.5\height}{\small{~(a) Sc-ResNet-50-NF}}
		&
		\raisebox{0.5\height}{\small{~(b) Sc-ResNet-50}}
		&
		\raisebox{0.5\height}{\small{~(c) Sc-ResNet-101}}
	\end{tabular}
	\caption{The Shannon entropy comparison of features from each block in the second stage between (a): our full Sc-ResNet-50 and Sc-ResNet-50 without flipping (NF),
	(b): our Sc-ResNet-50 and ResNet-50 and (c): our Sc-ResNet-101 and ResNet-101.}
	\label{se}
\end{figure}

Shannon entropy is a good choice to measure the diversity of features~\cite{chu2014analysis}, which is often adopted to calculate the uncertainty of the information contained in the data. Increasing the uncertainty of the information makes the shallow features richer and more interpretable. Here, to verify our SlimConv, we draw statistics for Shannon entropy of three blocks at the second stage which stands for the low-level information. As shown in Fig.~\ref{se}~(a), in general, our full Sc-ResNet-50 has larger Shannon entropy than the one without weights flipping, which means the flipping operation plays an important role in enhancing the diversity of features. Comparing with the plain ResNet-50 (Fig.~\ref{se}~(b)) and ResNet-101 (Fig.~\ref{se}~(c)), our models also shows larger Shannon entropy than the basic models. For this reason, our SlimConv module demonstrates the capability of improving the feature diversity while reducing redundancy. 

\section{Conclusion}\label{Sec:Conclusion}
In this paper we have designed a novel SlimConv module, an efficient architectural unit to decrease computational cost and model storage while improving performances of deep CNN models by reducing channel redundancies. The SlimConv consists of three steps, namely Reconstruct, Transform and Fuse. A weight flipping operation has been proposed which can largely improve the feature diversities. The extensive experiments on multiple challenging tasks have shown the effectiveness of our SlimConv. The existing state-of-the-art methods that integrated with SlimConv not only reduce the computations and save the storage, but also possess performance improvements. In addition, the discussion section has indicated that the SlimConv equipped models have potentials for the further compression. Finally, we hope our proposed method can inspire the research for more efficient architectural design.

%
%
\bibliographystyle{splncs04}
\bibliography{egbib}

\begin{thebibliography}{10}
\providecommand{\url}[1]{\texttt{#1}}
\providecommand{\urlprefix}{URL }
\providecommand{\doi}[1]{https://doi.org/#1}

\bibitem{chen2018big}
Chen, C.F., Fan, Q., Mallinar, N., Sercu, T., Feris, R.: Big-little net: An
  efficient multi-scale feature representation for visual and speech
  recognition. arXiv preprint arXiv:1807.03848  (2018)

\bibitem{chen2018encoder}
Chen, L.C., Zhu, Y., Papandreou, G., Schroff, F., Adam, H.: Encoder-decoder
  with atrous separable convolution for semantic image segmentation. In: {Proc.
  ECCV}. pp. 801--818 (2018)

\bibitem{chen2019drop}
Chen, Y., Fan, H., Xu, B., Yan, Z., Kalantidis, Y., Rohrbach, M., Yan, S.,
  Feng, J.: Drop an octave: Reducing spatial redundancy in convolutional neural
  networks with octave convolution. In: {Proc. ICCV}. pp. 3435--3444 (2019)

\bibitem{chollet2017xception}
Chollet, F.: Xception: Deep learning with depthwise separable convolutions. In:
  {Proc. CVPR}. pp. 1251--1258 (2017)

\bibitem{chu2014analysis}
Chu, J.L., Krzy{\.z}ak, A.: Analysis of feature maps selection in supervised
  learning using convolutional neural networks. In: Canadian Conference on
  Artificial Intelligence. pp. 59--70. Springer (2014)

\bibitem{everingham2015pascal}
Everingham, M., Eslami, S.A., Van~Gool, L., Williams, C.K., Winn, J.,
  Zisserman, A.: The pascal visual object classes challenge: A retrospective.
  {International Journal of Computer Vision}  \textbf{111}(1),  98--136 (2015)

\bibitem{everingham2010pascal}
Everingham, M., Van~Gool, L., Williams, C.K., Winn, J., Zisserman, A.: The
  pascal visual object classes (voc) challenge. {International Journal of
  Computer Vision}  \textbf{88}(2),  303--338 (2010)

\bibitem{fan2019shifting}
Fan, D.P., Wang, W., Cheng, M.M., Shen, J.: Shifting more attention to video
  salient object detection. In: {Proc. CVPR}. pp. 8554--8564 (2019)

\bibitem{gao2019res2net}
Gao, S., Cheng, M.M., Zhao, K., Zhang, X.Y., Yang, M.H., Torr, P.H.: Res2net: A
  new multi-scale backbone architecture. {IEEE Trans. on Pattern Analysis and
  Machine Intelligence}  (2019)

\bibitem{han2016dsd}
Han, S., Pool, J., Narang, S., Mao, H., Gong, E., Tang, S., Elsen, E., Vajda,
  P., Paluri, M., Tran, J., et~al.: Dsd: Dense-sparse-dense training for deep
  neural networks. arXiv preprint arXiv:1607.04381  (2016)

\bibitem{han2015learning}
Han, S., Pool, J., Tran, J., Dally, W.: Learning both weights and connections
  for efficient neural network. In: {Proc. NIPS}. pp. 1135--1143 (2015)

\bibitem{hariharan2011semantic}
Hariharan, B., Arbel{\'a}ez, P., Bourdev, L., Maji, S., Malik, J.: Semantic
  contours from inverse detectors. In: {Proc. ICCV}. pp. 991--998 (2011)

\bibitem{he2016deep}
He, K., Zhang, X., Ren, S., Sun, J.: Deep residual learning for image
  recognition. In: {Proc. CVPR}. pp. 770--778 (2016)

\bibitem{he2017channel}
He, Y., Zhang, X., Sun, J.: Channel pruning for accelerating very deep neural
  networks. In: {Proc. CVPR}. pp. 1389--1397 (2017)

\bibitem{hu2018squeeze}
Hu, J., Shen, L., Sun, G.: Squeeze-and-excitation networks. In: {Proc. CVPR}.
  pp. 7132--7141 (2018)

\bibitem{huang2017densely}
Huang, G., Liu, Z., Van Der~Maaten, L., Weinberger, K.Q.: Densely connected
  convolutional networks. In: {Proc. CVPR}. pp. 4700--4708 (2017)

\bibitem{iandola2016squeezenet}
Iandola, F.N., Han, S., Moskewicz, M.W., Ashraf, K., Dally, W.J., Keutzer, K.:
  Squeezenet: Alexnet-level accuracy with 50x fewer parameters and< 0.5 mb
  model size. arXiv preprint arXiv:1602.07360  (2016)

\bibitem{ioffe2015batch}
Ioffe, S., Szegedy, C.: Batch normalization: Accelerating deep network training
  by reducing internal covariate shift. arXiv preprint arXiv:1502.03167  (2015)

\bibitem{ke2017multigrid}
Ke, T.W., Maire, M., Yu, S.X.: Multigrid neural architectures. In: {Proc.
  CVPR}. pp. 6665--6673 (2017)

\bibitem{krizhevsky2009learning}
Krizhevsky, A., Hinton, G., et~al.: Learning multiple layers of features from
  tiny images  (2009)

\bibitem{krizhevsky2012imagenet}
Krizhevsky, A., Sutskever, I., Hinton, G.E.: Imagenet classification with deep
  convolutional neural networks. In: {Proc. NIPS}. pp. 1097--1105 (2012)

\bibitem{li2016pruning}
Li, H., Kadav, A., Durdanovic, I., Samet, H., Graf, H.P.: Pruning filters for
  efficient convnets. arXiv preprint arXiv:1608.08710  (2016)

\bibitem{li2019selective}
Li, X., Wang, W., Hu, X., Yang, J.: Selective kernel networks. In: {Proc.
  CVPR}. pp. 510--519 (2019)

\bibitem{lin2014microsoft}
Lin, T.Y., Maire, M., Belongie, S., Hays, J., Perona, P., Ramanan, D.,
  Doll{\'a}r, P., Zitnick, C.L.: Microsoft coco: Common objects in context. In:
  {Proc. ECCV}. pp. 740--755 (2014)

\bibitem{liu2018progressive}
Liu, C., Zoph, B., Neumann, M., Shlens, J., Hua, W., Li, L.J., Fei-Fei, L.,
  Yuille, A., Huang, J., Murphy, K.: Progressive neural architecture search.
  In: {Proc. ECCV}. pp. 19--34 (2018)

\bibitem{luo2018thinet}
Luo, J.H., Zhang, H., Zhou, H.Y., Xie, C.W., Wu, J., Lin, W.: Thinet: pruning
  cnn filters for a thinner net. {IEEE Trans. on Pattern Analysis and Machine
  Intelligence}  \textbf{41}(10),  2525--2538 (2018)

\bibitem{ma2018shufflenet}
Ma, N., Zhang, X., Zheng, H.T., Sun, J.: Shufflenet v2: Practical guidelines
  for efficient cnn architecture design. In: Proceedings of the European
  Conference on Computer Vision (ECCV). pp. 116--131 (2018)

\bibitem{marrero2019feratt}
Marrero~Fernandez, P.D., Guerrero~Pena, F.A., Ren, T., Cunha, A.: Feratt:
  Facial expression recognition with attention net. In: {Proc. CVPRW}. pp.~0--0
  (2019)

\bibitem{nair2010rectified}
Nair, V., Hinton, G.E.: Rectified linear units improve restricted boltzmann
  machines. In: {Proc. ICML}. pp. 807--814 (2010)

\bibitem{paszke2019pytorch}
Paszke, A., Gross, S., Massa, F., Lerer, A., Bradbury, J., Chanan, G., Killeen,
  T., Lin, Z., Gimelshein, N., Antiga, L., et~al.: Pytorch: An imperative
  style, high-performance deep learning library. In: {Proc. NIPS}. pp.
  8024--8035 (2019)

\bibitem{qiu2019deeplidar}
Qiu, J., Cui, Z., Zhang, Y., Zhang, X., Liu, S., Zeng, B., Pollefeys, M.:
  Deeplidar: Deep surface normal guided depth prediction for outdoor scene from
  sparse lidar data and single color image. In: {Proc. CVPR}. pp. 3313--3322
  (2019)

\bibitem{rastegari2016xnor}
Rastegari, M., Ordonez, V., Redmon, J., Farhadi, A.: Xnor-net: Imagenet
  classification using binary convolutional neural networks. In: {Proc. ECCV}.
  pp. 525--542 (2016)

\bibitem{ren2015faster}
Ren, S., He, K., Girshick, R., Sun, J.: Faster r-cnn: Towards real-time object
  detection with region proposal networks. In: {Proc. NIPS}. pp. 91--99 (2015)

\bibitem{russakovsky2015imagenet}
Russakovsky, O., Deng, J., Su, H., Krause, J., Satheesh, S., Ma, S., Huang, Z.,
  Karpathy, A., Khosla, A., Bernstein, M., et~al.: Imagenet large scale visual
  recognition challenge. {International Journal of Computer Vision}
  \textbf{115}(3),  211--252 (2015)

\bibitem{sandler2018mobilenetv2}
Sandler, M., Howard, A., Zhu, M., Zhmoginov, A., Chen, L.C.: Mobilenetv2:
  Inverted residuals and linear bottlenecks. In: {Proc. CVPR}. pp. 4510--4520
  (2018)

\bibitem{selvaraju2017grad}
Selvaraju, R.R., Cogswell, M., Das, A., Vedantam, R., Parikh, D., Batra, D.:
  Grad-cam: Visual explanations from deep networks via gradient-based
  localization. In: {Proc. ICCV}. pp. 618--626 (2017)

\bibitem{simonyan2014very}
Simonyan, K., Zisserman, A.: Very deep convolutional networks for large-scale
  image recognition. arXiv preprint arXiv:1409.1556  (2014)

\bibitem{szegedy2015going}
Szegedy, C., Liu, W., Jia, Y., Sermanet, P., Reed, S., Anguelov, D., Erhan, D.,
  Vanhoucke, V., Rabinovich, A.: Going deeper with convolutions. In: {Proc.
  CVPR}. pp.~1--9 (2015)

\bibitem{tan2019mnasnet}
Tan, M., Chen, B., Pang, R., Vasudevan, V., Sandler, M., Howard, A., Le, Q.V.:
  Mnasnet: Platform-aware neural architecture search for mobile. In: {Proc.
  CVPR}. pp. 2820--2828 (2019)

\bibitem{tan2019efficientnet}
Tan, M., Le, Q.V.: Efficientnet: Rethinking model scaling for convolutional
  neural networks. arXiv preprint arXiv:1905.11946  (2019)

\bibitem{tan2019efficientdet}
Tan, M., Pang, R., Le, Q.V.: Efficientdet: Scalable and efficient object
  detection. arXiv preprint arXiv:1911.09070  (2019)

\bibitem{wang2017residual}
Wang, F., Jiang, M., Qian, C., Yang, S., Li, C., Zhang, H., Wang, X., Tang, X.:
  Residual attention network for image classification. In: {Proc. CVPR}. pp.
  3156--3164 (2017)

\bibitem{woo2018cbam}
Woo, S., Park, J., Lee, J.Y., So~Kweon, I.: Cbam: Convolutional block attention
  module. In: {Proc. ECCV}. pp. 3--19 (2018)

\bibitem{xie2017aggregated}
Xie, S., Girshick, R., Doll{\'a}r, P., Tu, Z., He, K.: Aggregated residual
  transformations for deep neural networks. In: {Proc. CVPR}. pp. 1492--1500
  (2017)

\bibitem{yu2018deep}
Yu, F., Wang, D., Shelhamer, E., Darrell, T.: Deep layer aggregation. In:
  {Proc. CVPR}. pp. 2403--2412 (2018)

\bibitem{yu2019autoslim}
Yu, J., Huang, T.: Autoslim: Towards one-shot architecture search for channel
  numbers. arXiv preprint arXiv:1903.11728  \textbf{8} (2019)

\bibitem{yu2019universally}
Yu, J., Huang, T.S.: Universally slimmable networks and improved training
  techniques. In: {Proc. CVPR}. pp. 1803--1811 (2019)

\bibitem{yu2018slimmable}
Yu, J., Yang, L., Xu, N., Yang, J., Huang, T.: Slimmable neural networks. arXiv
  preprint arXiv:1812.08928  (2018)

\bibitem{zhang2018image}
Zhang, Y., Li, K., Li, K., Wang, L., Zhong, B., Fu, Y.: Image super-resolution
  using very deep residual channel attention networks. In: {Proc. ECCV}. pp.
  286--301 (2018)

\bibitem{zoph2018learning}
Zoph, B., Vasudevan, V., Shlens, J., Le, Q.V.: Learning transferable
  architectures for scalable image recognition. In: {Proc. CVPR}. pp.
  8697--8710 (2018)

\end{thebibliography}
\newpage
\section*{Appendix}
\appendix
\section{CIFAR}
To validate the performance of our SlimConv on small sized dataset, we conduct experiments on the CIFAR-100 dataset~\cite{krizhevsky2009learning}, which contains 50k images for the training and 10k images for the testing with 100 classes. The size of testing images are $32\times32$. We replace the corresponding layer of the basic block with our proposed SlimConv module. For fair comparisons, we keep the same training and testing strategy unchanged. 

\subsection{Comparing with lightweight models}
Table.~\ref{cp1} reports $3$ groups of results according to the complexity. When equipped with our SlimConv, the most widely used lightweight models (ShuffleNet~\cite{ma2018shufflenet} and MobileNet~\cite{sandler2018mobilenetv2}) both achieve over $1\%$ better accuracy with less computational cost and parameters. In particular, we further integrate the neural-architecture-search model (NasNet~\cite{zoph2018learning}) with our SlimConv module, yielding improved top-1 error as well as less consumption of resources than the original non-equipped basic model.

\begin{table}[H]
\begin{center}
\begin{tabular}{l|c|c|c}
\hline
Model & Top-1 Error & FLOPs($10^6$) & Params($10^6$) \\
\hline
\hline
1.0 ShuffleNet(v2) & 29.46 & 46.22 & 1.36\\
1.0 Sc-ShuffleNet(v2)(ours) & \textbf{27.74} & \textbf{44.58} & \textbf{1.34}\\
\hline
1.0 MobileNet(v2) & 28.15 & 94.72 & 2.41\\
1.0 Sc-MobileNet(v2)(ours) & \textbf{26.73} & \textbf{74.80} & \textbf{2.19}\\
\hline
NasNet & 20.85 & 681.88 & 5.22\\
Sc-NasNet(ours) & \textbf{20.56} & \textbf{644.71} & \textbf{5.05} \\
\hline
\end{tabular}
\end{center}
\caption{Performance comparison for ShuffleNet~\cite{ma2018shufflenet}, MobileNet~\cite{sandler2018mobilenetv2}, NasNet~\cite{zoph2018learning} and our integrated models on CIFAR-100.}
\label{cp1}
\end{table}

\subsection{Comparing with middle sized models}
We conduct $4$ groups of experiments for middle sized models to test our performances with different network design mechanisms. Table~\ref{cp2} reports the results. In the first group,  our integrated Sc-ResNet-50 achieves almost $1.4\%$ better accuracy, almost $36\%$ less FLOPs and $37\%$ less parameters than non-equipped original ResNet-50~\cite{he2016deep}. In the second group, we embed our SlimConv into the last stage of Oct-ResNet-50~\cite{chen2019drop}, which achieves almost $0.3\%$ better accuracy than the baseline while reducing about $60$ MFLOPs and $15.5\%$ parameters. When compared to the SOTA method SKNet-50~\cite{li2019selective} which takes ResNeXt-50~\cite{xie2017aggregated} as the basic model, our integrated Sc-ResNeXt-50 also achieves better accuracy, over $22\%$ less FLOPs and $26.7\%$ less parameters. Next, we study the impact of feature channel calibration by using SE-ResNet~\cite{hu2018squeeze} as the basic model. The SlimConv-equipped SE-ResNet-50 achieves about $0.6\%$ less top-1 error, $35.7\%$ less FLOPs and $33.5\%$ less parameters than the original model. As network layers increased, our integrated Sc-SE-ResNet-101 achieves almost $0.8\%$ better accuracy, $38.4\%$ computational cost and $34.9\%$ parameters compared with SE-ResNet-101. It is worth mentioning that Sc-ResNet-50 already has better performance than SE-ResNet-101.
\begin{table}[th]
\begin{center}
\begin{tabular}{l|c|c|c}
\hline
Model & Top-1 Error & FLOPs($10^9$) & Params($10^6$) \\
\hline
\hline
ResNet-50 & 22.88 & 1.305 & 23.71\\
Sc-ResNet-50(ours) & \textbf{21.51} & \textbf{0.836} & \textbf{14.91}\\
\hline
Oct-ResNet-50($\alpha = 0.5$) & 19.68 & 0.936 & 23.71\\
Sc-Oct-ResNet-50($\alpha = 0.5$)(ours) & \textbf{19.41} & \textbf{0.876} & \textbf{20.04}\\
\hline
SKNet-50 & 19.84 & 1.568 & 28.16\\
Sc-ResNeXt-50(ours) & \textbf{19.79} & \textbf{1.221} & \textbf{20.64}\\
\hline
SE-ResNet-50 & 22.26 & 1.317 & 26.50 \\
Sc-SE-ResNet-50(ours) & \textbf{21.63} & \textbf{0.847} & \textbf{17.61}\\
SE-ResNet-101 & 21.75 & 2.538 & 47.77 \\
Sc-SE-ResNet-101(ours) & \textbf{20.96} & \textbf{1.564} & \textbf{31.08}\\
\hline
\end{tabular}
\end{center}
\caption{Performance comparison for ResNet-50~\cite{he2016deep}, Oct-ResNet-50~\cite{chen2019drop}, SKNet-50~\cite{li2019selective}, SE-ResNet~\cite{hu2018squeeze}, ResNext~\cite{xie2017aggregated} and our integrated models on CIFAR-100.}
\label{cp2}
\vspace{-2.0em}
\end{table}

\section{Compressibility}
\begin{table}[th]
\begin{center}
\begin{tabular}{l|c|c|c|c|c|c}
\hline
Model & k & Top-1 Error & FLOPs($10^9$) & Params($10^6$) & Compressed($\%$) & \tabincell{c}{Latency\\($10^{-2}$ms)}\\
\hline
\hline
ResNet-50 & - & 22.88 & 1.305 & 23.71 & - & -\\
\hline
\hline
\multirow{7}{*}{Sc-ResNet-50} & $\sfrac{4}{3}$ & \textbf{21.51} & 0.836 & 14.91 & 37.12 & 5\\
\cline{2-7}
~& 2 & 22.39 & 0.653 & 11.56 & 51.24 & 3\\
\cline{2-7}
~& $\sfrac{8}{3}$ & 21.93 & 0.584 & 10.25 & 56.77 & 4\\
\cline{2-7}
~& $\sfrac{10}{3}$ & 23.83 & 0.542 & 9.52 & 59.85 & 4\\
\cline{2-7}
~& 4 & 24.15 & 0.519 & 9.08 & 61.70 & 4\\
\cline{2-7}
~& $\sfrac{14}{3}$ & 23.74 & 0.505 & 8.80 & 62.88 & 8\\
\cline{2-7}
~& $\sfrac{16}{3}$ & 24.56 & \textbf{0.495} & \textbf{8.61} & \textbf{63.69} & 6\\
\hline
\end{tabular}
\end{center}
\caption{Compressibility for our Sc-ResNet-50 with different hyperparameters(k) on CIFAR-100. The latency time is counted when the size of input image is $32\times32$.}
\label{cb}
\end{table}
Our SlimConv is a plug-and-play module, which can be easily integrated into CNNs to effectively compress models by only changing one hyperparameter($k$). We conduct several experiments with different values of $k$. The results are reported in Table~\ref{cb}. As the value of $k$ increases, the compression ratio also increases with slightly drops of the performances. As such, the value of $k$ is a tread-off value that can be tuned for different applications according to the computational resources. Specifically, our integrated Sc-ResNet-50 can still achieve nearly $1\%$ better accuracy than the basic model with $k$ set to $\frac{8}{3}$ while reducing parameters by $56.77\%$. In addition, we count the latency time (ms per image) for the integrated models on a single GPU. 

\end{document}